%% file: main.tex
\documentclass{article}

\usepackage{iclr2027_conference,times}
\usepackage{amsmath,amssymb}
\usepackage{booktabs}
\usepackage{float}
\usepackage{graphicx}
\usepackage{microtype}
\usepackage{multirow}
\usepackage{tabularx}
\usepackage[table]{xcolor}
\usepackage{enumitem}
\usepackage{pifont}
\usepackage{makecell}
\usepackage{hyperref}
\hypersetup{hidelinks,pdfauthor={Haitong Ma, Chenxiao Gao, Rushi Qiang, Bo Dai, Na Li},pdftitle={RLE-Bench: A Qualifying Exam for Coding Agents as Robot Learning Engineers},pdfsubject={},pdfkeywords={}}
\ifdefined\pdfinfoomitdate\pdfinfoomitdate=1\fi
\ifdefined\pdftrailerid\pdftrailerid{}\fi
\ifdefined\pdfsuppressptexinfo\pdfsuppressptexinfo=15\fi
\usepackage{cleveref}
\usepackage{url}
\usepackage{tcolorbox}
\usepackage{wrapfig}

\input{content/task_cards/style}

\definecolor{tbdred}{RGB}{180,25,25}
\definecolor{todoorange}{RGB}{190,90,0}

\newcommand{\rlebench}{\textsc{RLE-Bench}}

\newcommand{\task}[1]{$\mathtt{Family#1}$}

\definecolor{BenchGreen}{HTML}{1BAF7A}
\definecolor{BenchRed}{HTML}{E34948}
\definecolor{BenchBlue}{HTML}{2A78D6}

\newcommand{\benchyes}{\textcolor{BenchGreen}{\ding{51}}}
\newcommand{\benchno}{\textcolor{BenchRed}{\ding{55}}}
\newcolumntype{Y}{>{\centering\arraybackslash}X}

\title{RLE-Bench: A Qualifying Exam for Coding Agents as Robot Learning Engineers}

\author{Haitong Ma$^{1,*}$ \quad Chenxiao Gao$^{2,*}$ \quad Rushi Qiang$^{2,*}$ \quad Bo Dai$^{2,\dagger}$ \quad Na Li$^{1,\dagger}$ \\[4pt]
        \normalfont $^{1}$Harvard University \quad $^{2}$Georgia Institute of Technology \\[4pt]
        \normalfont\small $^{*}$Equal contribution \quad $^{\dagger}$Equal supervision \\[4pt]
        \normalfont Project website: \href{https://rle-bench.github.io/}{\textcolor{BenchBlue}{\nolinkurl{https://rle-bench.github.io/}}}}

\iclrfinalcopy
\renewcommand{\headrulewidth}{0.4pt}
\renewcommand{\footrulewidth}{0pt}

\begin{document}

\maketitle
\fancyhead{}
\fancyhead[L]{\small\scshape RLE-Bench: A Qualifying Exam for Coding Agents}
\fancyhead[R]{\small Preprint}
\fancyfoot{}
\fancyfoot[C]{\thepage}

\input{content/abstract}
\input{content/introduction}

\input{content/related_work}
\input{content/rle_bench}

\input{content/results}
\input{content/conclusion}

\bibliography{references}
\bibliographystyle{iclr2027_conference}

\newpage
\appendix
\input{content/appendix_task_cards}
\input{content/appendix_full_results}

\end{document}

%% file: content/task_cards/style.tex
\definecolor{cardline}{HTML}{BCC3CA}
\definecolor{cardhead}{HTML}{EEF1F4}
\definecolor{cardcream}{HTML}{FFF6E4}
\definecolor{cardgray}{HTML}{F5F5F7}
\definecolor{cardblue}{HTML}{EAF5F7}
\definecolor{cardgreen}{HTML}{EDF5E9}
\definecolor{cardrose}{HTML}{FBEDEA}
\newcommand{\cardpanel}[3]{%
  \begingroup\setlength{\fboxsep}{7pt}%
  \colorbox{#1}{\begin{minipage}{\dimexpr\linewidth-14pt\relax}%
  \raggedright
  \textbf{\MakeUppercase{#2}}\par\vspace{3pt}#3%
  \end{minipage}}\endgroup\par\vspace{5pt}}
\newenvironment{carditems}{\begin{itemize}[leftmargin=10pt,itemsep=2pt,topsep=0pt,parsep=0pt]}{\end{itemize}}

%% file: content/abstract.tex
\begin{abstract}

Coding agents are beginning to move beyond purely digital tasks to tackle physical-world challenges, particularly in robotics. Existing robotics benchmarks, however, primarily focus on the performance of individual artifacts, such as policies or controllers, offering limited coverage of coding agents’ broader engineering capabilities. Real-world robotics extends beyond control: agents must build, integrate, diagnose, and improve heterogeneous artifacts under resource constraints and reason from multimodal feedback. To evaluate these broader capabilities, we introduce \rlebench{}, a benchmark of robot-learning tasks spanning four representative robotics development workflows: interactive control, policy learning, perception and estimation, and mechanical design. We use diverse task-specific metrics to evaluate the artifacts submitted by the coding agents, from the success rate the agents achieved to the policy agents trained, the harness agent built, and the mechanical structures the agent designed. We aggregate these metrics into an overall RLE Index and report workflow-specific capability profiles, enabling systematic comparison of coding agents' capabilities across multiple capability dimensions. Beyond performance ranks, we also conduct in-depth case studies examining agent behavior on representative tasks, highlighting both current capabilities and limitations, and pointing to the opportunities robotics tasks have to offer for future agent training.

\end{abstract}

\begin{figure}[h]
    \centering
    \includegraphics[width=0.85\linewidth]{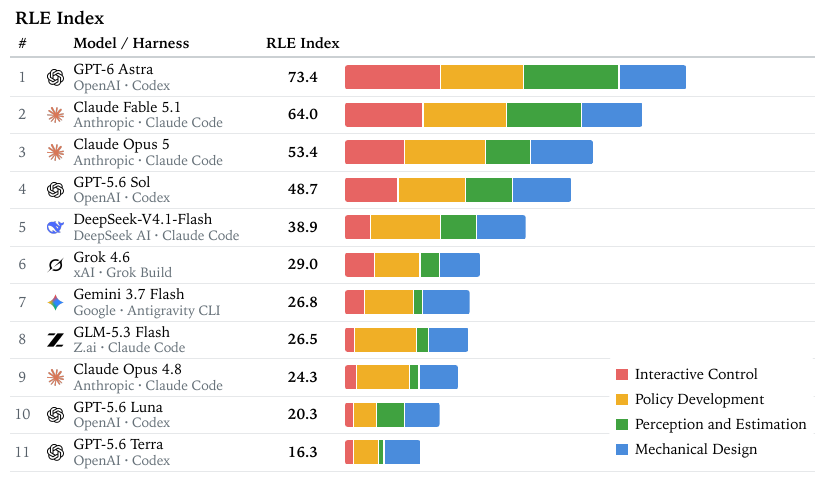}
    \caption{\textbf{\rlebench{} Leaderboard.}
Models are ranked by the overall RLE Index. 
}
\label{fig:main_leaderboard}
\end{figure}

%% file: content/introduction.tex
\section{Introduction}
\label{sec:introduction}

Coding agents have moved well beyond code completion. They can now inspect unfamiliar repositories, write and execute programs, interpret test failures, and revise artifacts over long trajectories. Benchmarks have played a key role in this transition, both by defining the need and by providing signals for how to scale up training. For example, repository-level coding benchmarks such as SWE-bench~\citep{jimenez2024swe} and interactive environments such as Terminal-Bench~\citep{merrill2026terminal} have made this transition measurable, while machine-learning engineering benchmarks~\citep{huang2023mlagentbench,chan2025mle,qiang2026mle} further test their capability in training, evaluating, and improving machine learning systems at a larger scale.  

Beyond digital engineering tasks, interacting with and reasoning about the physical world remains the grand challenge for LLM agents. Unlike purely digital settings, robotics requires agents to understand spatiotemporal relationships, reason about the physical consequences of their decisions, and adapt their actions based on multimodal feedback from the environment. Robotics engineering, which covers diverse topics such as mechanical design, hardware-software calibration, closed-loop control, failure diagnosis, and policy development, provides an excellent testbed to showcase these physically grounded capabilities. However, existing robotics benchmarks~\citep{liu2023libero,mu2025robotwin,chen2026robodojo} primarily evaluate policies manifested as machine learning models, such as vision-language-action models (VLAs,~\cite{brohan2022rt,black2024pi_0}) and world-action models (WAMs,~\cite{ye2026world}), rather than the broader capabilities coding agents exercise throughout the robotics development process. As an early step toward evaluating coding agents in robotics, CaP-X~\citep{fu2026capx} systematically benchmarks an agent's ability to write code as controllers for manipulation tasks across different levels of abstraction; however, its evaluation remains limited to writing code as policies rather than the holistic robot-learning workflow.


To fill this gap and measure whether frontier agents can transfer their capabilities into real-world physical tasks, 
we introduce \rlebench{}, which asks: 
\begin{tcolorbox}[
    colback=blue!5!white,
    colframe=blue!60!black,
    boxrule=0.7pt,
    arc=2mm,
    left=3mm, right=3mm,
    top=2mm, bottom=2mm
]
\centering\itshape
Can general-purpose coding agents qualify for the diverse,
iterative, and physically grounded work of robot-learning
engineers?
\end{tcolorbox}
The benchmark comprises nine families of tasks organized into four primary workflow families: \textit{Interactive Control}, \textit{Policy Learning},  \textit{Perception and Estimation}, and \textit{Mechanical Design}. Each task provides an instruction and an interactive simulator through which the agent can observe, act, and receive physical feedback. The resulting agentic systems or executable artifacts are evaluated using benchmark-owned instrumentation under hidden scenes, dynamics, embodiments, random seeds, or held-out tasks.

A key distinction of \rlebench{} from existing robotics benchmarks is its emphasis on a broader range of robot-learning workflows: rather than evaluating policies, controllers, or other components in isolation, \rlebench{} evaluates whether coding agents can perform and improve the heterogeneous tasks that collectively constitute robotics development. This holistic perspective is important for the future we envision, in which coding agents can organically integrate capabilities across the development stack and use them to iteratively improve both their robotic systems and the infrastructure used to build them.

We evaluate 11 model–harness combinations and report the \textbf{RLE Index}, an equally weighted macro-average of the four workflow-family scores in the benchmark, as our primary metric for each combination. Beyond this scalar metric, we provide per-workflow profiles that illustrate the performance and cost trade-offs across the capabilities we care about for robotics. We also conduct in-depth case studies examining agent behavior on representative tasks, highlighting both current capabilities and limitations, and pointing to the opportunities robotics tasks offer for future agent training.

%% file: content/related_work.tex
\section{Related Work}
\label{sec:related}

\paragraph{Software Engineering (SWE) and Machine Learning Engineering (MLE) Benchmarks.}
To faithfully evaluate agents' coding capabilities and provide interactive testbeds in which they can act and learn from feedback, a growing number of coding-oriented benchmarks have emerged. Among them, SWE-Bench \citep{jimenez2024swe} curates real-world repositories and issues from the web and tasks agents with generating code patches, which are evaluated against functional tests to provide feedback on their actions. OSWorld evaluates agents in interactive computer environments using task-specific execution-based checks~\citep{xie2024osworld}, while Terminal-Bench evaluates long-horizon terminal tasks in isolated execution environments~\citep{merrill2026terminal}. Beyond traditional SWE tasks, recent work has increasingly focused on machine learning engineering. MLAgentBench and MLE-Bench shift the focus from software repair to experimental machine learning, requiring agents to train, evaluate, and improve machine learning models~\citep{huang2023mlagentbench,chan2025mle}. MLE-Dojo provides a Gym-style environment spanning over 200 Kaggle challenges, with structured interaction, online outcome verification, and support for both agent evaluation and training~\citep{qiang2026mle}. \rlebench{} builds on this paradigm of executable, iterative evaluation while extending it to robot learning, where agent actions are grounded in physical consequences and directly interact with physical simulators. 

\paragraph{Robotics Policy Evaluation Benchmarks.}
Numerous benchmarks have been developed to facilitate the prototyping and evaluation of robot policies. These benchmarks typically provide physics simulators, task environments, and evaluation protocols, and can be broadly categorized by task domain. RLBench~\citep{james2020rlbench}, CALVIN~\citep{mees2022calvin}, LIBERO~\citep{liu2023libero}, RoboSuite~\citep{zhu2020robosuite}, RoboCasa~\citep{nasiriany2024robocasa}, RoboTwin~\citep{mu2025robotwin,chen2025robotwin}, and RoboDojo~\citep{chen2026robodojo} focus primarily on manipulation with fixed-base or mobile manipulators. Gym-Locomotion~\citep{brockman2016openai} and DMControl~\citep{tassa2018deepmind} cover locomotion, while broader platforms such as IsaacLab~\citep{mittal2025isaac} and ManiSkill~\citep{tao2024maniskill3} support manipulation, locomotion, and whole-body control. However, these benchmarks primarily evaluate individual artifacts, such as controller or VLA policies, rather than the broader capabilities of coding agents in performing robotics development workflows.

\paragraph{Coding Agents for Robotics.}
A predominant use of coding agents in robotics is to generate controller code for robot control. Code as Policies~\citep{liang2023code} and VoxPoser~\citep{huang2023voxposer} demonstrate that language-model-generated programs can solve manipulation tasks. Building on this paradigm, CaP-X systematically evaluates code-as-policy agents across varying levels of abstraction, interaction, and perceptual grounding~\citep{fu2026capx}. Concurrent work explores coding agents for improving policy repositories, reproducing robot-learning systems, and operating physical experimentation loops~\citep{elmaaroufi2026rho,xiao2026enpire,jin2026nautilus}. Beyond controller generation, LLM agents have also been used to generate reward functions for reinforcement learning, as exemplified by Text2Reward~\citep{xie2024text2reward} and Eureka \citep{ma2024eureka}. In contrast to these specialized applications, \rlebench{} takes a holistic perspective, systematically evaluating coding agents across the broader robot-development stack.

\Cref{tab:benchmark-comparison} compares \rlebench{} against existing benchmarks along each dimension.


\begin{table}[t]
\centering
\begingroup
\fontsize{7}{8.5}\selectfont
\setlength{\tabcolsep}{3pt}
\renewcommand{\arraystretch}{1.25}

\caption{
Comparison of benchmark evaluation scope.
The SWE / MLE group includes SWE-bench, Terminal-Bench,
MLE-bench, and MLE-Dojo.
}\label{tab:benchmark-comparison}
\begin{tabularx}{\linewidth}{@{}l|YY|YYYY@{}}
\toprule
\textbf{Benchmark}
&
\makecell{\textbf{Coding}\\\textbf{Agents}}
&
\makecell{\textbf{Physics}\\\textbf{Grounded}}
&
\makecell{\textbf{Controller}\\\textbf{Synthesis}}
&
\makecell{\textbf{Robot Policy}\\\textbf{Training}}
&
\makecell{\textbf{Perception and}\\\textbf{Estimation}}
&
\makecell{\textbf{Mechanical}\\\textbf{Design}}
\\
\midrule

\makecell[l]{SWE/MLE benchmarks}
& \benchyes & \benchno
& \benchno & \benchno & \benchno & \benchno
\\

\makecell[l]{LIBERO/RoboTwin/RoboDojo}
& \benchno & \benchyes
& \benchno & \benchyes & \benchno & \benchno
\\

CaP-X
& \benchyes & \benchyes
& \benchyes & \benchno & \benchno & \benchno
\\

\rowcolor{BenchBlue!10}
\textbf{RLE-Bench}
& \benchyes & \benchyes
& \benchyes & \benchyes & \benchyes & \benchyes
\\

\bottomrule
\end{tabularx}
\endgroup
\end{table}

%% file: content/rle_bench.tex
\section{\rlebench{}}
\label{sec:benchmark}
\begin{figure}[t]
    \centering
\includegraphics[width=\linewidth]{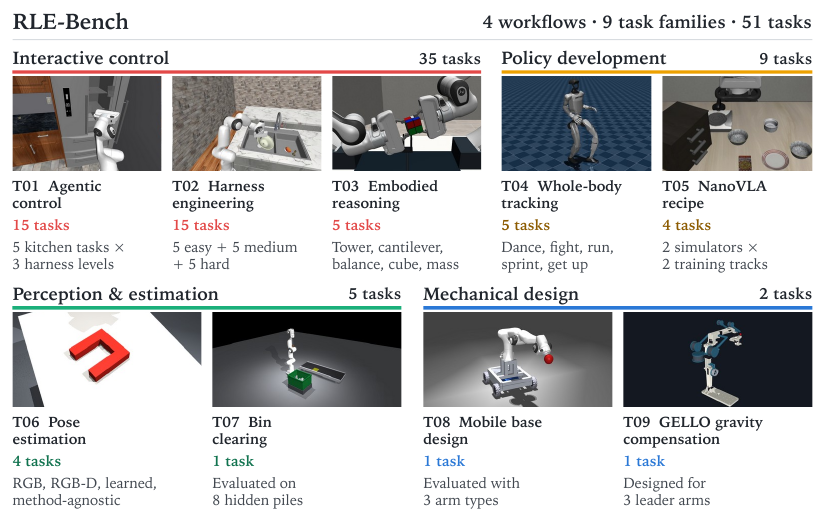}
    \caption{Overview of the \rlebench{}.}
    \label{fig:main_demo}
\end{figure}

In this section, we detail the design of \rlebench{}, covering its task templates, budget control, design principles of each workflow, and the evaluation protocol. \Cref{fig:main_demo} provides an overview of the tasks in \rlebench{}.

\begin{table}[h]
\caption{The \rlebench{} task categories. A total of 51 tasks are categorized into 4 workflows, 9 families. Each family shares a similar capability requirement, artifact contract, and the evaluation metrics.}
\label{tab:tasks}
\centering
\scriptsize
\resizebox{\linewidth}{!}{%
\begin{tabular}{cllll}
\toprule
Task Family & Workflow & Objective & Submitted artifact & Metrics\\
\midrule
\task{01} & Interactive control & RoboCasa task learning & Agent context & Success rate \\
\task{02} & Interactive control & Reusable RoboCasa harness & Harness code + manual & Success rate \\
\task{03} & Interactive control & Tabletop physical reasoning & -- & Solution optimality \\
\task{04} & Policy learning & Whole-body motion tracking & Model checkpoint & Motion tracking error \\
\task{05} & Policy learning & VLA recipe engineering & Model checkpoint & Validation loss \\
\task{06} & Perception \& estimation & Blind multi-shape pose estimation & Pose estimator code / models & Estimation error \\
\task{07} & Perception \& estimation & Contact-rich bin clearing & End-to-end controller code & Operation efficiency \\
\task{08} & Mechanical design & Common mobile-manipulator base design & MJCF design + controller code & Design metric \\
\task{09} & Mechanical design & GELLO gravity compensation module design & MJCF design + controller code & Design metric \\
\bottomrule
\end{tabular}}
\end{table}

\subsection{Task Description and Evaluation Setup}

A benchmark task in \rlebench{} is a tuple
\begin{equation}
\tau = (I, E_{\mathrm{dev}}, B, \mathcal{A}, E_{\mathrm{eval}}, M),
\end{equation}
where $I$ is the task instruction in plain text, $E_{\mathrm{dev}}$ is the public development environment such as a scratch workspace with connections to a RoboCasa simulator~\citep{nasiriany2024robocasa}, $B$ is the budget including total wall clock time, interaction steps, compute resources, and etc.  $\mathcal{A}$ is an executable artifact contract, $E_{\mathrm{eval}}$ is a verifier-controlled evaluation environment, and $M$ is a verifier to evaluate the artifact submitted by the agent. 

\rlebench{} has 51 tasks in total.
We organize them into four primary workflows according to the focused capabilities: \textit{Interactive Control} (\task{01-03}), \textit{Policy Development} (\task{04-05}), \textit{Perception and Estimation} (\task{06-07}), and \textit{Mechanical Design} (\task{08-09}). They are further categorized into 9 families according to the actual task objective, submitted artifacts, and evaluation metrics, which is summarized in~\Cref{tab:tasks}. A detailed walkthrough of each task is provided in \Cref{sec:task_family}.

Each task in the suite is split into a development phase and an evaluation phase. In the development phase, the agent interacts with environment $E_{\rm dev}$ under resource constraints $B$ and builds and submits an artifact $a$. In the evaluation phase, the verifier $M$ evaluates the submitted artifact $a$ in the evaluation environment $E_{\rm eval}$. To receive a valid score, the artifact is generally required to satisfy the contract specified by the instruction $I$.

We construct \rlebench{} from two complementary sources. The first category, comprising \task{01,02,04,05}, draws on simulators, datasets, and task definitions from existing open-source benchmarks and projects, which we repurpose under new objectives to enable a standardized, comprehensive evaluation of coding agents. The second category, comprising \task{03,06,07,08,09}, is hand-designed by us, drawing inspiration from daily robotics engineering practice.

\subsection{A Walkthrough of the Workflows and Task Families}
\label{sec:task_family}

In this section, we introduce the details about each task and its design motivations.

\paragraph{Workflow 1: Interactive Control (\task{01-03}).} 
Before an agent can build anything for a robot, it has to be able to operate one. Direct interaction is exactly where physical grounding is most exposed, since observations are partial, actions cannot be undone, and progress depends on reading multimodal feedback correctly. In these tasks the agent interacts with the world directly to achieve given goals, either by controlling the robot itself, by building the tools that let another agent do so, or by acting to gather the evidence a decision needs. 

Specifically, \task{01} evaluates agents to perform closed-loop control on 5 RoboCasa \citep{nasiriany2024robocasa} tasks under three different levels of harness that progressively add reusable utilities and privileged simulator state. \task{02} instead asks agents to package their experience into reusable tools and lessons, which will be used by fresh agents to solve held-out unseen tasks. \task{03} focuses on embodied reasoning, where agents must interact with the environment and gather information to infer task-critical physical properties, rather than relying on static visual QA. 

\paragraph{Workflow 2: Policy Learning (\task{04-05}).} 
Direct control does not scale to complicated tasks such as dexterous manipulations and high-frequency locomotion. In such scenarios, we must resort to a learned policy or controller, and the training recipes, such as the dataset, reward function, curricula, and model architecture, are the key to the final policy performance. Besides, policy learning is also another way for the agent to turn solved instances and experiences into durable, reusable capability. 

\task{04} tackles locomotion, asking the agent to train humanoid whole-body controllers through motion tracking. The agent is provided with motion clips from the LAFAN1 dataset \citep{harvey2020robust} and must set up the entire training pipeline to train the tracking policy, which is then evaluated Sim2Sim under hidden variations in dynamics, latency, and sensor noise.
\task{05} is named \emph{nanoVLA}, inspired by the nanoGPT project for rapid iteration on training smaller-scale language models. We select LIBERO~\citep{liu2023libero} and RoboTwin~\citep{chen2025robotwin} and their paired datasets, ask the agent to develop the training pipeline and VLA architecture under resource constraints, and evaluate the delivered model on a selected subset of benchmark tasks. 


\paragraph{Workflow 3: Perception \& Estimation (\task{06-07}).}
Sensors are central to robotic systems, and coding agents must understand them well to make good decisions. Real perception systems are noisy, frequently occluded, and rate-limited, so these tasks directly evaluate how well agents understand the physical world through sensor readings. 

\task{06} consists of four 2D pose estimation problems in a tabletop block-pushing task, where the block can move and become occluded by the robot arm, creating significant difficulty for pose estimation and requiring agents to reason about when occlusion occurs and how to handle it; to further constrain the scope, we define four variants in which agents are supplied with different sensors and inference hardware configurations. \task{07} asks agents to write a code-as-policy controller for a simplified industrial pick-and-place task, where a manipulator equipped with two RGB-D cameras, force-torque sensors, and a magnetic gripper must pick \emph{all} U-shaped metal components from a bin onto a conveyor belt as fast as possible.



\textbf{Workflow 4: Mechanical Design (\task{08-09}).} 
Hardware bounds what any controller can achieve. An under-powered base or an uncompensated arm cannot be fixed in software, so an engineer has to reason about mass, torque, geometry, and build-ability together with the control that will run on them. This workflow evaluates mechanical design and its supporting control software as a coupled system.

\task{08} asks agents to develop a common mobile-manipulator base and controller for multiple robot arms including Franka Research 3, XArm7 and UR5e. The agents need to achieve the goal of reaching all the reference points on a shelf, while balancing payload, resource use, and stability. \task{09} requires adding gravity compensation to GELLO lead-arms~\citep{wu2024gello}, evaluated under unseen physical instances, poses, and payloads. Both tasks make mechanical design and its supporting software part of the solution, rather than treating the embodiment as fixed.

\subsection{Orchestration and Evaluation Protocol}

To support standardized evaluation and integration, all tasks in \rlebench{} are shipped as Harbor environments~\citep{team2026harbor}. Each agent receives the task instruction, a declared artifact contract, public assets, tools, and a resource-bounded development session. Once development ends, the artifact crosses into the verifier image, where the hidden seeds, privileged states, and reference assets all remain verifier-private to prevent information leakage. The verifier loads the artifact in a sandboxed or constrained subprocess, executes physical rollouts, and produces the only authoritative reward report. 


%% file: content/results.tex
\section{Evaluation Results}
\label{sec:results}

\subsection{Setup and Metrics}

\begin{wraptable}{r}{0.55\linewidth}
    \centering
    \small
    \caption{Evaluated model-harness combinations.
    }
    \label{tab:evaluated-systems}
    \resizebox{1.0\linewidth}{!}{%
    \begin{tabular}{@{}ll@{}}
        \toprule
        \textbf{Model} & \textbf{Harness} \\
        \midrule
        Claude Fable 5.1~\citep{anthropic2026fable51}
            & Claude Code \\
        Claude Opus 5~\citep{anthropic2026opus5}
            & Claude Code \\
        Claude Opus 4.8~\citep{anthropic2026opus48}
            & Claude Code \\
        GPT-6 Astra~\citep{openai2026gpt6astra}
            & Codex CLI \\
        GPT-5.6 Sol~\citep{openai2026gpt56}
            & Codex CLI \\
        GPT-5.6 Terra~\citep{openai2026gpt56}
            & Codex CLI \\
        GPT-5.6 Luna~\citep{openai2026gpt56}
            & Codex CLI \\
        Gemini 3.7 Flash~\citep{google2026gemini37flash}
            & Antigravity CLI \\
        Grok 4.6~\citep{grok46}
            & Grok Build \\
        GLM-5.3-Flash~\citep{zai2026glm53flash}
            & Claude Code \\
        DeepSeek-V4.1-Flash~\citep{deepseek2026v41flashcard}
            & Claude Code \\
        \bottomrule
    \end{tabular}
    }
\end{wraptable}

\paragraph{Agents and Models}
\label{sec:agents-models}
We evaluate the 11 model--harness combinations listed in
Table~\ref{tab:evaluated-systems}.
The coding-agent harness provides agents with the tools to read/write files, execute bash commands, and manage contexts, and the task-specific robotics scaffoldings described above are added on top of these harnesses. 
Each evaluation configuration records the exact model identifier,
harness version, reasoning setting, and resource budget. For Claude series, GPT series, Gemini series, and Grok series models, we use the default context management strategy configured by their own harness; for open-sourced models, including GLM-5.3-Flash and DeepSeek-V4.1-Flash, we use the same context management strategy as Claude Opus 5. Reasoning efforts are set to \textit{high} for all combinations. Web-search tools are banned by default to prevent information leakage. 

\paragraph{Metrics}

Each task $\tau_i$ from family $\mathcal{F}_j$ and workflow $\mathcal{G}_k$ retains a native score $S_{i, j, k}\in [0,1]$.
We aggregate the score at three levels: {average score across subtasks} $\to$ {average score across families within one workflow} $\to$ {average score across workflows}. The final score is what we report as \textbf{RLE-Index}. 
Formally, for a primary family $\mathcal{F}_j$ and workflow $\mathcal{G}_k$, we define
\begin{equation}
S_{j,k}=\frac{1}{|\mathcal{F}_{j}|}\sum_{\tau_i\in\mathcal{F}_{j}}S_{i,j,k},
\qquad
S_{k}=\frac{1}{|\mathcal{G}_{k}|}\sum_{f_j\in\mathcal{G}_{k}}S_{j,k},
\qquad
S^{\mathrm{RLE}}=\frac{1}{4}\sum_{k=1}^{4}S_{k}.
\label{eq:rlescore}
\end{equation}
The overall leaderboard is ranked by $S^{\mathrm{RLE}}$. Thus, each workflow family has equal weight. We show family scores in the same table and report all native task metrics in the appendix. 



\subsection{Leaderboard, Capability Profiles and Cost}

\begin{figure}
    \centering
    \includegraphics[width=\linewidth]{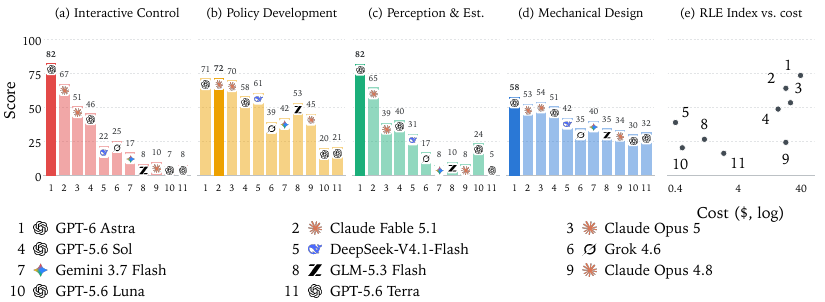}
    \caption{Capability profile per workflow (a-d) and performance versus cost (e).}
    \label{fig:score_per_workflow_and_costs}
\end{figure}

The overall \rlebench{} leaderboard is shown in~\Cref{fig:main_leaderboard}, with per-workflow scores compared in~\Cref{fig:score_per_workflow_and_costs}.

\textbf{Finding \#1: GPT-6 Astra and Claude Fable 5.1 lead the leaderboard, with visual grounding emerging as a key differentiator.}
GPT-6 Astra and Claude Fable 5.1 outperform the other open- and closed-source models by a substantial margin. Their advantage is most pronounced in \textit{Interactive Control} and \textit{Perception and Estimation}, both of which require interpreting multimodal observations and acting on repeated environment feedback. This pattern suggests that stronger visual grounding contributes substantially to their overall lead.

\textbf{Finding \#2: Performance is more closely matched in policy development.}
In \textit{Policy Development}, which more closely resembles traditional machine learning engineering tasks, the performance gap between models is substantially smaller.

\textbf{Finding \#3: Understanding physical consequences remains hard for all models. }On tasks from \textit{Mechanical Design}, performance gaps are much smaller, and no model consistently makes sound design decisions, particularly when implementation choices have delayed or indirect physical consequences.

\subsection{Case Studies and Key Insights}

\begin{figure}[h]
    \centering
    \includegraphics[width=\linewidth]{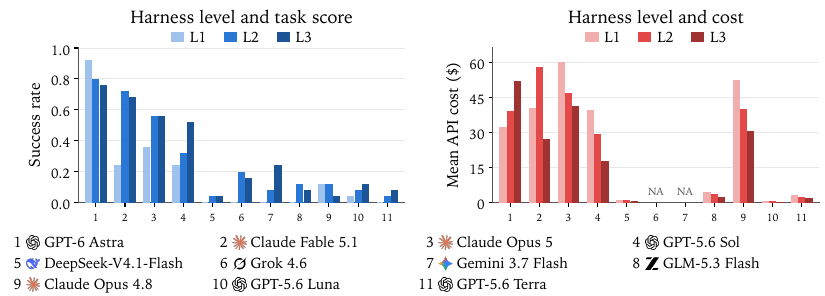}
    \caption{Effect of harness level on task success rate and cost.}
    \label{fig:harness_level}
\end{figure}

\textbf{Finding \#4: External robotics scaffolding generally helps, until the model is strong enough. }In \task{01}, we evaluate three scaffolding levels: \texttt{L1} provides basic robot and sensor APIs; \texttt{L2}, inspired by CaP-X~\citep{fu2026capx}, adds SAM3~\citep{carion2026sam} and Contact-GraspNet~\citep{sundermeyer2021contact} for perception and grasp planning; and \texttt{L3} adds privileged object and fixture positions. Figure~\ref{fig:harness_level} shows that richer scaffolding generally improves task scores and reduces costs for most models, including GPT-5.6, the Opus series, and Gemini-3.7-Flash. GPT-6-Astra, however, performs strongly with \texttt{L1}, gaining little or occasionally losing performance with additional support. These results suggest that external modules become less beneficial as model capabilities improve and may sometimes interfere with the model's own perception and reasoning.

\begin{figure}[t]
    \centering
    \includegraphics[width=\linewidth]{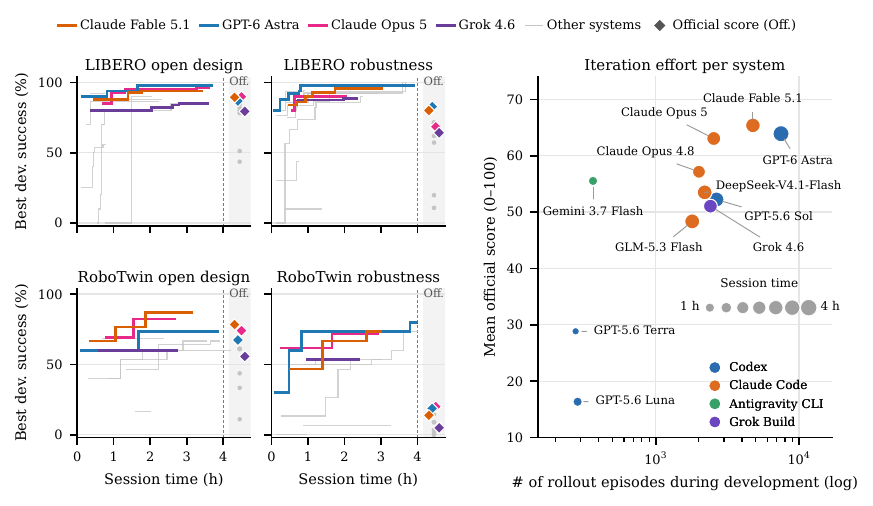}
    \caption{Development progress in the four \task{05} subtasks. \textbf{Left:} best development success so far over the four-hour session, one line per session; lines end when the agent stops, and markers in the shaded column represents the final evaluation score. \textbf{Right:} rollout episodes each system requested during development against its mean official score.}
    \label{fig:policy_self_improvement}
\end{figure}

\textbf{Finding \#5: Evaluation feedback drives real improvements within the scope it measures. }
In \task{05}, agents may query an evaluation service to obtain a preliminary assessment of their policy during development. We monitor and log this process, which allows us to replay the agents' progress in \Cref{fig:policy_self_improvement}. Inspecting the curves from the development stage, we find that the development-time evaluation service delivers real, front-loaded improvement. In 29 of 36 sessions, the submitted version outperforms the agent's first development-time evaluation. From the right figure, we also found that the evaluation score tends to correlate positively with the rollout episodes requested during development. However, comparing final development scores against final official scores reveals that this improvement is bounded by what the feedback can measure. On open-design subtasks, where the evaluation task set is rehearsable, development success predicts the official score within 5 points. On robustness subtasks, however, whose test-time perturbations are never shown to the agent, the official score falls 19 (LIBERO) and 41 (RoboTwin) points below the development score. Several agents construct their own perturbation proxies, but none closes the gap, indicating that building robust policies remains a difficult challenge.
\label{sec:self_improvement}

\begin{figure}[!htb]
    \centering
    \includegraphics[width=1.0\linewidth]{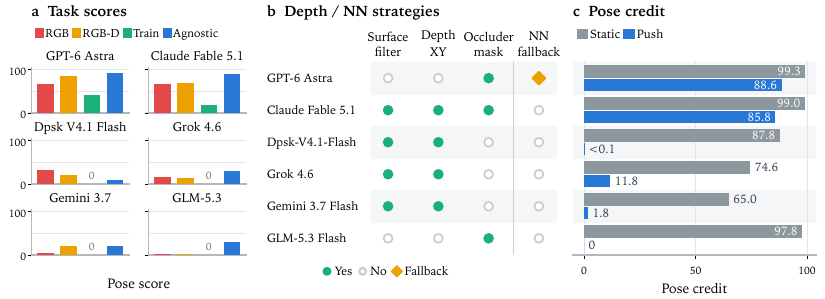}
    \caption{Behavior analysis in \task{06}, pose estimation. \textbf{Left:} scores of selected agents in different tasks in \task{06}; \textbf{Middle:} how the depth observation is used in method-agnostic task; \textbf{Right: }Score comparison between static frames and dynamic, occluded frames in method-agnostic task.}
    \label{fig:depth_behavior_analysis}
\end{figure}

\textbf{Finding \#6: Depth does improve perception, while multi-sensor fusion remains difficult. }\Cref{fig:depth_behavior_analysis} analyzed how the depth sensor affect perception system performance in \task{06}. First, most models improve the perception score with RGB-D information compared to RGB only, showing preliminary capability to understand multiple types of physical sensors. The real gap appears when we compare different strategies for using depth information with occlusion. High-performing models use depth to judge the binary occlusion masks rather than only recover the surface or 3D positions. Notably, GPT-6 Astra trains a small convolutional neural network as a fallback when depth information is not reliable, making it the only agent that actually uses the GPU in the method-agnostic task. Final results showed that, even for tasks such as simple 2D pose estimation, only the flagship models can build a reliable perception system when occlusions get in the way.

\begin{figure}[h]
    \centering
    \includegraphics[width=0.9\linewidth]{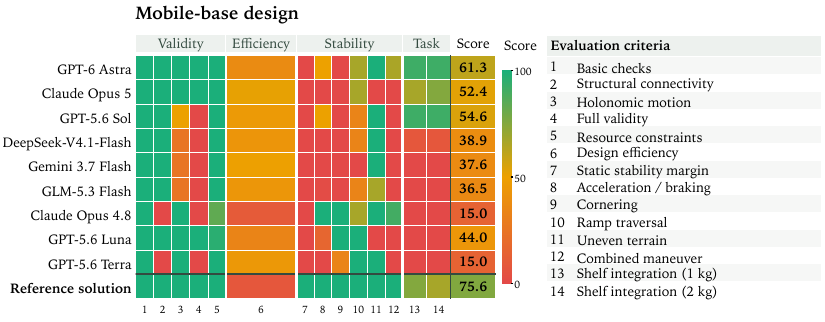}
    \caption{Sub-score of selected verifier evaluation criteria in mobile base design task results. }
\label{fig:mechanical_design_analysis}
\end{figure}

\textbf{Finding \#7: Agents can satisfy visible objectives while still missing coupled physical consequences. }
\Cref{fig:mechanical_design_analysis} presents selected results from the \task{08} mobile base design task. Designs are evaluated for structural validity, design efficiency, static and dynamic stability, and task completion. The instructions indicate all required properties without revealing the underlying verification protocols. Most models pass the structural validity checks, demonstrating their ability to produce mechanically plausible designs. However, these designs often fail physical evaluation: all nine models receive zero worst-arm credit for static stability margin. These cases show that agents can satisfy visible design objectives while overlooking coupled physical consequences, particularly instability under load and collisions during motion.

%% file: content/conclusion.tex
\section{Conclusion}
We introduced \rlebench{} as a qualifying exam for coding agents as robot learning engineers through executable, physics-grounded tasks spanning four representative workflow families. The benchmark combines an overall leaderboard with capability profiles, trusted hidden evaluation, and explicit resource accounting. We hope \rlebench{} provides a reproducible instrument for developing agents that can build, test, and improve reliable robot-learning systems.

Although this benchmark evaluates agents exclusively in simulation, errors in agent-generated code, learned policies, or mechanical designs could have serious physical consequences if deployed in the real world. We therefore encourage responsible use of the benchmark, human oversight when interpreting its results, and independent validation before deployment.

The benchmark also has several limitations. First, simulation performance does not establish real-world reliability or safety. The scale and diversity of the simulated scenes are insufficient to assess agent behavior across all deployment conditions. Second, the four representative workflows covered in this version capture only part of the work performed by robot learning engineers; real-world robotics requires a broader range of capabilities. Third, the planned public release introduces a risk of future training data contamination, which could compromise the validity of subsequent evaluations.

%% file: content/appendix_task_cards.tex
\clearpage
\section{Complete Task Cards}
\label{app:tasks}

\refstepcounter{subsection}\label{sec:task01}
\addcontentsline{toc}{subsection}{\protect\numberline{\thesubsection}Task 01: RoboCasa Speed-Run}
\input{content/task_cards/task01}

\clearpage
\refstepcounter{subsection}\label{sec:task02}
\addcontentsline{toc}{subsection}{\protect\numberline{\thesubsection}Task 02: RoboCasa Harness Transfer}
\input{content/task_cards/task02}

\clearpage
\refstepcounter{subsection}\label{sec:task03}
\addcontentsline{toc}{subsection}{\protect\numberline{\thesubsection}Task 03: Tabletop Physical Reasoning}
\input{content/task_cards/task03}

\clearpage
\refstepcounter{subsection}\label{sec:task04}
\addcontentsline{toc}{subsection}{\protect\numberline{\thesubsection}Task 04: Whole-Body Motion Tracking}
\input{content/task_cards/task04}

\clearpage
\refstepcounter{subsection}\label{sec:task05}
\addcontentsline{toc}{subsection}{\protect\numberline{\thesubsection}Task 05: nanoVLA Recipe Engineering}
\input{content/task_cards/task05}

\clearpage
\refstepcounter{subsection}\label{sec:task06}
\addcontentsline{toc}{subsection}{\protect\numberline{\thesubsection}Task 06: Blind Planar Pose Estimation}
\input{content/task_cards/task06}

\clearpage
\refstepcounter{subsection}\label{sec:task07}
\addcontentsline{toc}{subsection}{\protect\numberline{\thesubsection}Task 07: Contact-Rich Bin Clearing}
\input{content/task_cards/task07}

\clearpage
\refstepcounter{subsection}\label{sec:task08}
\addcontentsline{toc}{subsection}{\protect\numberline{\thesubsection}Task 08: Universal Mobile-Manipulator Base}
\input{content/task_cards/task08}

\clearpage
\refstepcounter{subsection}\label{sec:task09}
\addcontentsline{toc}{subsection}{\protect\numberline{\thesubsection}Task 09: Multi-Robot GELLO Co-Design}
\input{content/task_cards/task09}

\clearpage

%% file: content/task_cards/task01.tex
\begingroup
\hypersetup{hidelinks}
\setlength{\parindent}{0pt}
\setlength{\parskip}{0pt}
\fontsize{9}{10.8}\selectfont
\setlength{\fboxsep}{9pt}
\setlength{\fboxrule}{0.5pt}
\fcolorbox{cardline}{white}{%
\begin{minipage}{\dimexpr\linewidth-19pt\relax}
\cardpanel{cardhead}{}{\vspace{-9pt}%
{\large\bfseries Task Family 01: RoboCasa Speed-Run}\hfill{\small\bfseries 15 instances}}
\textbf{Family:} Interactive control\hfill\textbf{Environment:} RoboCasa\par
\textbf{Task matrix:} Five kitchen tasks $\times$ three harness levels\par
\vspace{6pt}
\cardpanel{cardcream}{Agent Instructions (partial)}{%
You are controlling a simulated Franka-class arm on a mobile base in a RoboCasa kitchen. The task is \textbf{OpenFridge}---open the refrigerator door. Your goal in this phase is to \textbf{learn to solve it and write the code you will need}. [\ldots] Evaluation will use the same task in unseen kitchen layouts.}
\cardpanel{cardgray}{Environment / execution view}{%
{\centering\includegraphics[width=.85\linewidth]{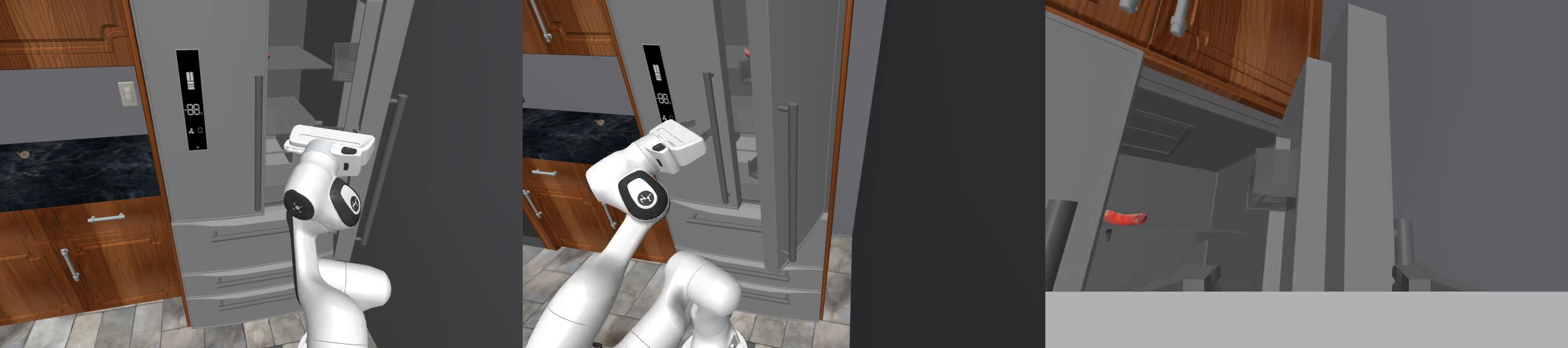}\par}
\vspace{3pt}
\textbf{Example: OpenFridge.} Three camera views from the \href{https://robocasa.ai/docs/build/html/tasks/atomic_tasks.html}{official RoboCasa demonstration} (illustrative).\par
\vspace{3pt}
Develop on \texttt{pretrain} kitchens, then evaluate in five distinct \texttt{target} kitchens. Files persist; conversation resumes when supported.\par
\vspace{3pt}
\textbf{Five task variants}\par\vspace{2pt}
\begin{tabularx}{\linewidth}{@{}l@{\hspace{8pt}}X@{}}
\texttt{OpenFridge} & Open the refrigerator door.\\
\texttt{CloseCabinet} & Close the cabinet door.\\
\texttt{TurnOnStove} & Turn on the specified burner.\\
\texttt{PickPlaceCounterToDrawer} & Move the target into the drawer.\\
\texttt{PickPlaceMicrowaveToCounter} & Move the target to its counter destination.
\end{tabularx}}
\begin{minipage}[t]{0.49\linewidth}
\vspace{0pt}
\cardpanel{cardblue}{Provided harness}{%
\begin{carditems}
\item Task instruction, simulator client, and a persistent coding workspace.
\item \textbf{L1:} Three RGB cameras, optional metric depth, and robot proprioception.
\item \textbf{L2 adds:} Camera geometry, frame transforms, point-cloud utilities, motion primitives, SAM3 segmentation, and Contact-GraspNet grasp proposals.
\item \textbf{L3 adds:} World-frame object and fixture poses, fixture extents, target identity, and grasp-related state.
\end{carditems}}
\cardpanel{cardblue}{Submitted artifacts}{%
\begin{carditems}
\item Controller and utility code developed in the workspace and used during evaluation.
\item Closed-loop 12-D actions in $[-1,1]$: arm (6), gripper (1), mobile base (3), torso (1), and mode (1).
\item Physical task completion, confirmed by the simulator's success predicate.
\end{carditems}}
\end{minipage}\hfill
\begin{minipage}[t]{0.49\linewidth}
\vspace{0pt}
\cardpanel{cardgreen}{Evaluation rubric}{%
Let $s$ be successful trials divided by five, and $d$ be development steps used:
\[
R=s\!\left[0.80+0.20\!\left(1-\frac{d}{50{,}000}\right)\right].
\]
\vspace{-7pt}
\begin{carditems}
\item Success determines 80\% of reward; interaction efficiency contributes up to 20\%, scaled by success.
\item Missing trials count as failures. A missing, unsealed, or unverifiable ledger scores zero.
\end{carditems}}
\cardpanel{cardrose}{Budget / evaluation protocol}{%
\begin{carditems}
\item \textbf{Develop:} 8 hours, 1 GPU, and 50,000 charged steps. Each reset and executed action costs one step; observations are free.
\item \textbf{Evaluate:} One terminal phase, 1 hour total; five trials with at most 1,000 actions each.
\item Evaluation actions do not enter $d$. Reset advances to the next trial.
\item Library-generated actions are metered, and the trusted ledger records interaction and success.
\end{carditems}}
\end{minipage}
\end{minipage}}
\endgroup

%% file: content/task_cards/task02.tex
\begingroup
\hypersetup{hidelinks}
\setlength{\parindent}{0pt}\setlength{\parskip}{0pt}
\fontsize{9}{10.8}\selectfont
\setlength{\fboxsep}{9pt}\setlength{\fboxrule}{0.5pt}
\fcolorbox{cardline}{white}{%
\begin{minipage}{\dimexpr\linewidth-19pt\relax}
\cardpanel{cardhead}{}{%
\vspace{-9pt}{\large\bfseries Task Family 02: RoboCasa Harness Transfer}\hfill{\small\bfseries 15 instances}}
\textbf{Family:} Interactive control\hfill\textbf{Environment:} RoboCasa\par
\textbf{Task matrix:} Five easy, five medium, and five hard transfer groups\par\vspace{6pt}
\cardpanel{cardcream}{Agent Instructions (partial)}{%
You are controlling a simulated Franka-class arm on a mobile base in a RoboCasa kitchen. [\ldots] Your job is a \textbf{harness}: perception primitives, controllers built on them, and a manual. When this phase ends, a \textbf{different agent} [\ldots] gets one shot at one kitchen task using only what you left behind. Your score is entirely what that agent achieves.}
\cardpanel{cardgray}{Environment / execution view}{%
{\centering\includegraphics[width=\linewidth,height=1.28in,keepaspectratio]{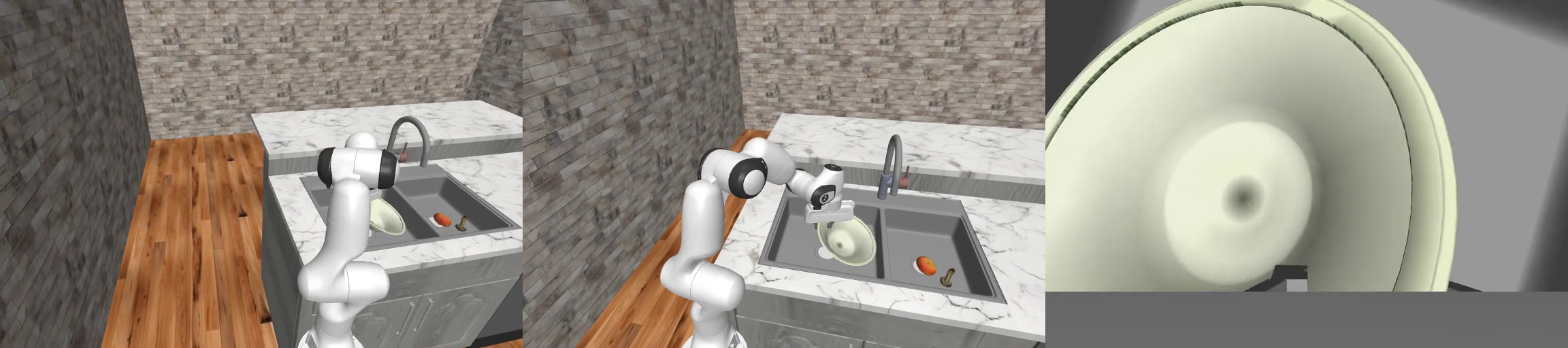}\par}\vspace{3pt}\textbf{Example: Washing Dishes.} The held-out task is \texttt{DumpLeftovers}, illustrated by an \href{https://robocasa.ai/docs/build/html/tasks/composite_tasks.html}{official RoboCasa demo}.\par\vspace{3pt}\begin{tabularx}{\linewidth}{@{}l@{\hspace{8pt}}X@{}}
Develop on & \texttt{PlaceOnDishRack}, \texttt{SortingCleanup}, \texttt{StackBowlsInSink}.\\
Transfer & Only \texttt{agent\_harness/} code and manual reach each fresh agent.\\
Evaluate & Five fresh agents independently attempt \texttt{DumpLeftovers}.\\
\end{tabularx}}
\begin{minipage}[t]{0.49\linewidth}\vspace{0pt}
\cardpanel{cardblue}{Provided harness}{%
\begin{carditems}
\item Three selected composite tasks per activity group; their atomic skills cover the held-out task.
\item Three RGB views, optional metric depth, robot proprioception, and a metered simulator client.
\item Observation-only perception: object and fixture poses are withheld.
\end{carditems}}
\cardpanel{cardblue}{Submitted artifacts}{%
\begin{carditems}
\item A reusable \texttt{agent\_harness/} with perception functions, closed-loop controllers, and a usage manual.
\item No development conversation or other workspace files are transferred.
\item Evaluation agents use the harness to execute 12-D robot actions.
\end{carditems}}
\end{minipage}\hfill\begin{minipage}[t]{0.49\linewidth}\vspace{0pt}
\cardpanel{cardgreen}{Evaluation rubric}{%
For baseline, peak simultaneous, and total satisfied conjunct counts $b,m,n$:\[s_{\rm trial}=\operatorname{clip}_{[0,1]}\!\left(\frac{m-b}{n-b}\right).\]\begin{carditems}
\item Simulator success gives 1. Otherwise, missing stage records or $n=b$ give 0.
\item Reward is the mean over five agents; missing trials count as zero.
\item Binary success is reported separately. Development efficiency adds no reward.
\end{carditems}}
\cardpanel{cardrose}{Budget / evaluation protocol}{%
\begin{carditems}
\item \textbf{Develop:} 8 h, 1 GPU, 75,000 charged steps; resets and executed actions each cost one step.
\item \textbf{Evaluate:} Five fresh agents, each with 1 h and 5,000 steps.
\item Observations are free. Evaluation uses unseen kitchens and the held-out task.
\end{carditems}}
\end{minipage}\end{minipage}}\endgroup

%% file: content/task_cards/task03.tex
\begingroup
\hypersetup{hidelinks}
\setlength{\parindent}{0pt}\setlength{\parskip}{0pt}
\fontsize{9}{10.8}\selectfont
\setlength{\fboxsep}{9pt}\setlength{\fboxrule}{0.5pt}
\fcolorbox{cardline}{white}{%
\begin{minipage}{\dimexpr\linewidth-19pt\relax}
\cardpanel{cardhead}{}{%
\vspace{-9pt}{\large\bfseries Task Family 03: Tabletop Physical Reasoning}\hfill{\small\bfseries 5 instances}}
\textbf{Family:} Interactive control\hfill\textbf{Environment:} MuJoCo / robosuite\par
\textbf{Task matrix:} Construction, physical inference, and contact manipulation\par\vspace{6pt}
\cardpanel{cardcream}{Agent Instructions (partial)}{%
Write a controller that solves a scrambled 64 mm $2\times2\times2$ Rubik\textquotesingle s Cube using two fixed Panda arms. [\ldots] Grasp opposing halves and rotate one half relative to the other. Turn layers through gripper contact and reorient the cube as needed to inspect hidden faces. Finish with all six faces solved and aligned, and the cube held still above the table.}
\cardpanel{cardgray}{Environment / execution view}{%
\begin{minipage}{.58\linewidth}\raggedright
{\centering\includegraphics[width=\linewidth,height=1.65in,keepaspectratio,trim=32bp 82bp 528bp 158bp,clip]{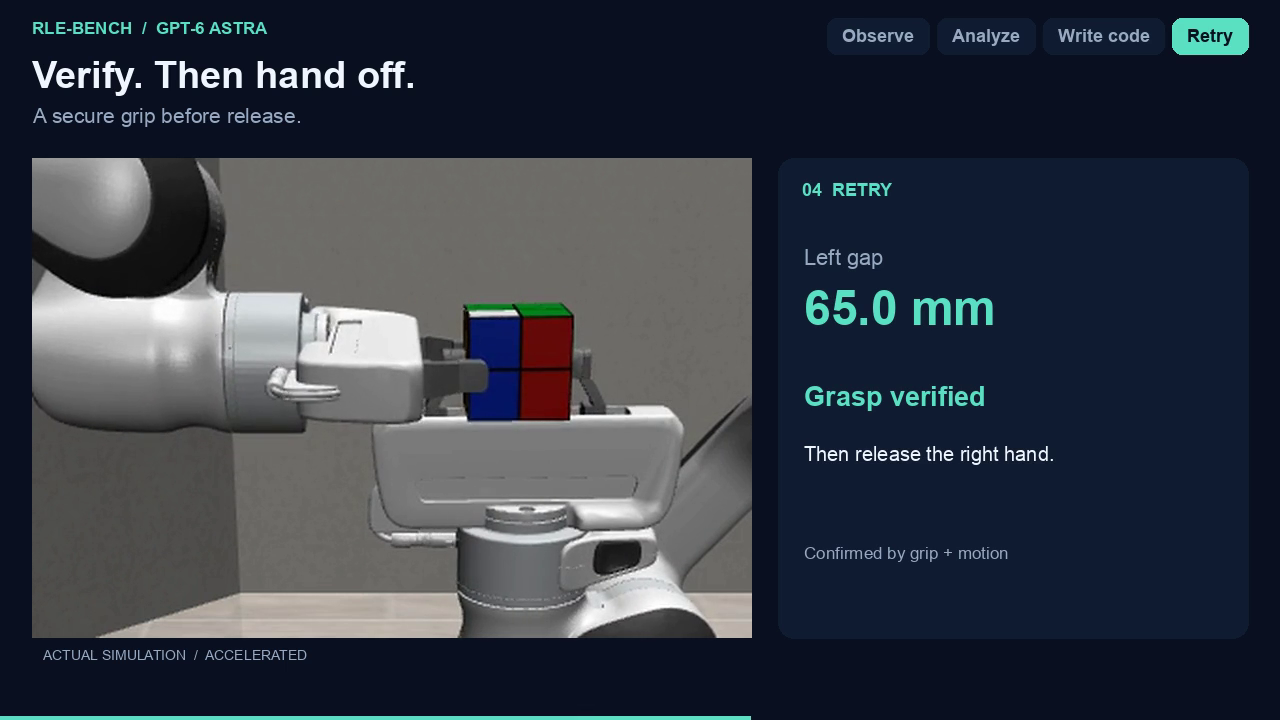}\par}
\end{minipage}\hfill\begin{minipage}{.39\linewidth}\raggedright
\textbf{Example: Pocket cube.} Two Panda grippers exchange the cube during a development attempt. Frame from a GPT-6 Astra evaluation trajectory, showing a retry after a missed handoff.
\end{minipage}\par\vspace{4pt}\begin{tabularx}{\linewidth}{@{}l@{\hspace{8pt}}X@{}}
Tower & Build a stable, unsupported tower from 12 mixed pieces.\\
Cantilever & Maximize the overhang of four supported blocks.\\
Balance & Identify the heavy cube among nine and place it on the mat.\\
Pocket cube & Solve a physical $2\times2\times2$ cube with two arms.\\
Hidden COM & Identify the ballast quadrant in each of three sealed boxes.\\
\end{tabularx}}
\begin{minipage}[t]{0.49\linewidth}\vspace{0pt}
\cardpanel{cardblue}{Provided harness}{%
\begin{carditems}
\item Tower / cantilever / balance: three RGB views, proprioception, 12-D actions, and a Cartesian movement helper.
\item Balance also exposes cube and pan positions; masses remain hidden.
\item Cube: one front RGB camera, optional depth, two-arm state, and 14-D actions.
\item Hidden COM: three RGB views and robot state; no force, contact, mass, or object-pose readings.
\end{carditems}}
\cardpanel{cardblue}{Submitted artifacts}{%
\begin{carditems}
\item Tower / cantilever / balance: the final physical scene.
\item Cube: a simulator-confirmed physical solve, with efficient face turns.
\item Hidden COM: one irreversible A/B/C/D answer per box; no correctness feedback.
\end{carditems}}
\end{minipage}\hfill\begin{minipage}[t]{0.49\linewidth}\vspace{0pt}
\cardpanel{cardgreen}{Evaluation rubric}{%
\begin{carditems}
\item Tower / cantilever: settled height or overhang divided by the analytic optimum, clipped to $[0,1]$.
\item Balance: zero unless only the heavy cube reaches the mat; then 1 for $w\leq2$ weighings, or $2/w$.
\item Cube: solve required; excess quarter turns reduce reward relative to the optimum. Invalid counters give zero; unclassified transitions cap reward at 0.5.
\item Hidden COM: fraction of three answers correct.
\end{carditems}}
\cardpanel{cardrose}{Budget / evaluation protocol}{%
\begin{carditems}
\item Tower / cantilever / balance / cube: 9 h, 1 GPU, 50,000 steps; one continuous attempt.
\item Resets restore the same scene for the first three tasks. Cube reset abandons the attempt; metered drop rescue preserves completed turns.
\item Hidden COM: 1 h total, 12,000 steps per box. Observations are free.
\end{carditems}}
\end{minipage}\end{minipage}}\endgroup

%% file: content/task_cards/task04.tex
\begingroup
\hypersetup{hidelinks}
\setlength{\parindent}{0pt}\setlength{\parskip}{0pt}
\fontsize{9}{10.8}\selectfont
\setlength{\fboxsep}{9pt}\setlength{\fboxrule}{0.5pt}
\fcolorbox{cardline}{white}{%
\begin{minipage}{\dimexpr\linewidth-19pt\relax}
\cardpanel{cardhead}{}{%
\vspace{-9pt}{\large\bfseries Task Family 04: Whole-Body Motion Tracking}\hfill{\small\bfseries 5 instances}}
\textbf{Family:} Policy development\hfill\textbf{Environment:} MuJoCo-Warp / MuJoCo-C\par
\textbf{Task matrix:} Dance, fight, fall and get up, run, sprint\par\vspace{6pt}
\cardpanel{cardcream}{Agent Instructions (partial)}{%
A Unitree G1 has to reproduce a 20-second excerpt of LAFAN1 \texttt{dance1\_subject2}: 29 joints, 50 Hz, matching the reference whole-body motion and world-frame trajectory without falling over. \textbf{The environment and a reference PPO are given. What you train is your choice.} [\ldots] Stage early, stage often.}
\cardpanel{cardgray}{Environment / execution view}{%
\begin{minipage}{.57\linewidth}\raggedright {\centering\includegraphics[width=\linewidth,height=1.6in,keepaspectratio,trim=450bp 80bp 450bp 180bp,clip]{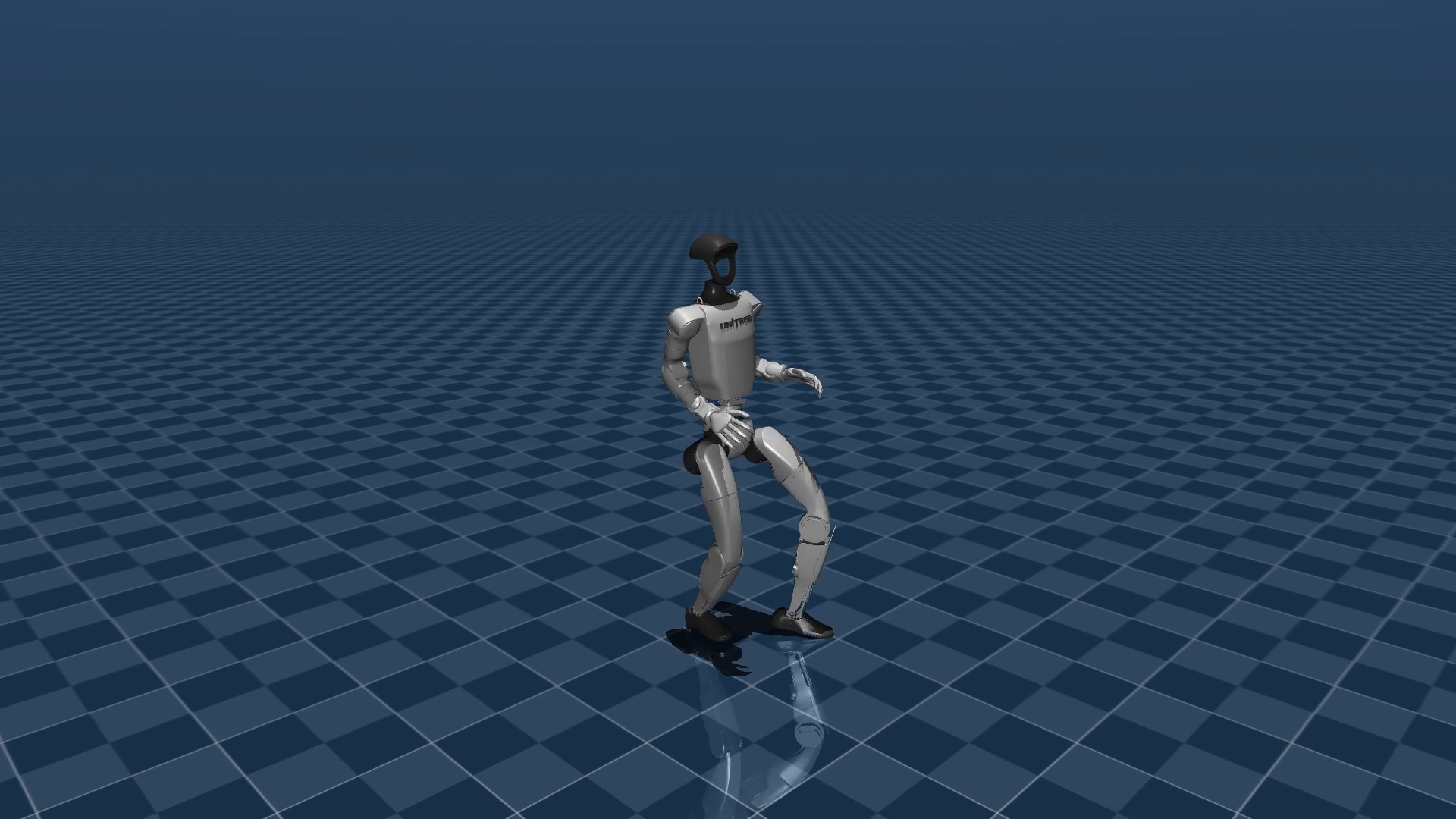}\par}\end{minipage}\hfill\begin{minipage}{.4\linewidth}\raggedright \textbf{Example: Dance tracking.} Unitree G1 executing a learned motion-tracking policy in simulation. Frame from a GPT-6 Astra evaluation trajectory.\par\vspace{5pt}\textbf{Workflow:} train in GPU-parallel MuJoCo-Warp, export ONNX, then evaluate in CPU MuJoCo-C.\end{minipage}\par\vspace{4pt}\begin{tabularx}{\linewidth}{@{}l@{\hspace{8pt}}X@{}}
Shared target & A 20-second reference clip for each of five motions.\\
Control interface & 160-D observation $\rightarrow$ 29 joint-position offsets at 50 Hz.\\
\end{tabularx}}
\begin{minipage}[t]{0.49\linewidth}\vspace{0pt}
\cardpanel{cardblue}{Provided harness}{%
\begin{carditems}
\item Retargeted LAFAN1 motion, G1 model, modifiable training environment, export tools, and design-seed evaluation.
\item An optional one-hour PPO baseline.
\item Observations include reference motion, torso-frame anchor, robot state, and previous action.
\end{carditems}}
\cardpanel{cardblue}{Submitted artifacts}{%
\begin{carditems}
\item \texttt{policy.onnx} and \texttt{policy\_meta.json}.
\item At most 10 million parameters; a history of 1--16 observation frames.
\item The verifier runs the exported graph directly; no agent training code is imported.
\end{carditems}}
\end{minipage}\hfill\begin{minipage}[t]{0.49\linewidth}\vspace{0pt}
\cardpanel{cardgreen}{Evaluation rubric}{%
Per hidden seed, score is clipped to $[0,1]$:\[\begin{aligned}s={}&0.7T+0.3U\\&-0.15\min(J/J_{\max},1)-0.05F.\end{aligned}\]\begin{carditems}
\item $T$: whole-body tracking; $U$: survival fraction; $J$: action jerk; $F$: fall indicator.
\item Reward is the mean over eight hidden seeds. Invalid exports score zero; stability or determinism smoke failure caps reward at 0.1.
\item Dance thresholds are calibrated; the other four sets remain provisional.
\end{carditems}}
\cardpanel{cardrose}{Budget / evaluation protocol}{%
\begin{carditems}
\item \textbf{Develop:} 4 h, 1 GPU; training and evaluation share the session budget.
\item Train at a 5 ms physics step; evaluate at 2 ms.
\item Hidden evaluation perturbs dynamics, sensors, terrain, latency, and pushes.
\end{carditems}}
\end{minipage}\end{minipage}}\endgroup

%% file: content/task_cards/task05.tex
\begingroup
\hypersetup{hidelinks}
\setlength{\parindent}{0pt}\setlength{\parskip}{0pt}
\fontsize{9}{10.8}\selectfont
\setlength{\fboxsep}{9pt}\setlength{\fboxrule}{0.5pt}
\fcolorbox{cardline}{white}{%
\begin{minipage}{\dimexpr\linewidth-19pt\relax}
\cardpanel{cardhead}{}{%
\vspace{-9pt}{\large\bfseries Task Family 05: nanoVLA Recipe Engineering}\hfill{\small\bfseries 4 instances}}
\textbf{Family:} Policy development\hfill\textbf{Environment:} LIBERO / RoboTwin 2.0\par
\textbf{Task matrix:} Two environments $\times$ open-design and robustness tracks\par\vspace{6pt}
\cardpanel{cardcream}{Agent Instructions (partial)}{%
Maximise LIBERO-10 success with DINOv2-base as the only pretrained component. You get demonstrations, the encoder and a socket protocol; you decide everything else: whether to freeze, fine-tune or partially reuse the encoder, how language is handled, the policy architecture, the optimiser, the serving code. [\ldots] Both modes run offline. Keep every stochastic operation seeded.}
\cardpanel{cardgray}{Environment / execution view}{%
{\centering\includegraphics[width=\linewidth,height=1.48in,keepaspectratio,trim=0bp 4096bp 2048bp 0bp,clip]{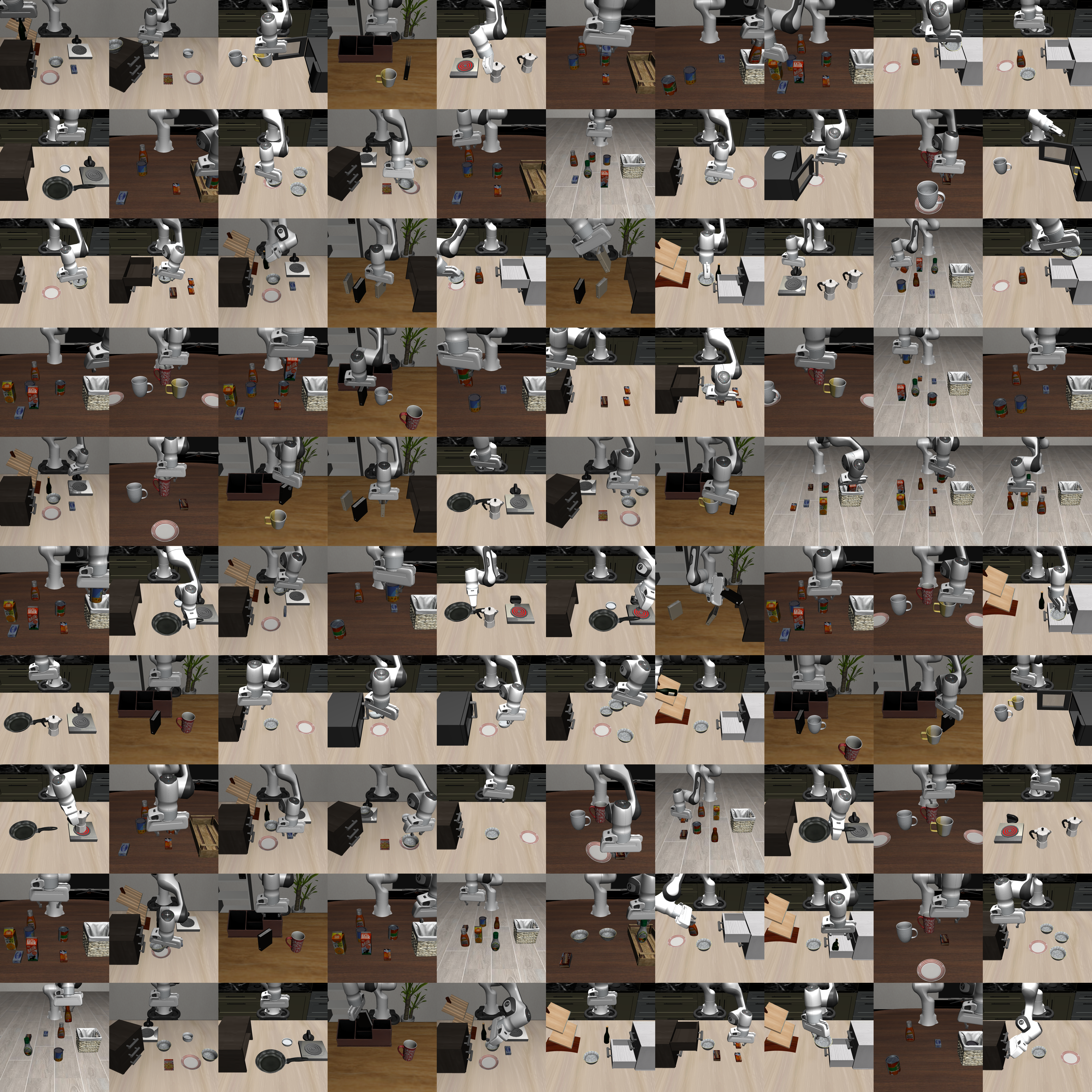}\par}\vspace{3pt}\textbf{Illustration:} Representative LIBERO scenes, cropped from the \href{https://libero-project.github.io/main.html}{official project collage}. RLE-Bench uses LIBERO-10 and RoboTwin tracks. Only the submitted Python recipe is transferred and retrained.\par\vspace{3pt}\begin{tabularx}{\linewidth}{@{}l@{\hspace{8pt}}X@{}}
LIBERO open / robust & 500 evaluation episodes each; at least 475 complete.\\
RoboTwin open / robust & 300 / 225 episodes; at least 285 / 214 complete.\\
\end{tabularx}}
\begin{minipage}[t]{0.49\linewidth}\vspace{0pt}
\cardpanel{cardblue}{Provided harness}{%
\begin{carditems}
\item Read-only demonstrations, approved pretrained encoders and loaders, a policy socket protocol, and public evaluation client.
\item Open-design tracks allow DINOv2-base; robustness tracks provide a vision/text encoder bundle.
\item No policy, trainer, simulator, or evaluator implementation is supplied.
\end{carditems}}
\cardpanel{cardblue}{Submitted artifacts}{%
\begin{carditems}
\item One UTF-8 \texttt{solution.py}, at most 1 MiB, supporting training and serving.
\item Replay must write \texttt{ckpt.pt}; \texttt{meta.json} is optional. Development checkpoints are not transferred.
\item Inference consumes language, two $128\times128$ RGB views, and robot state; returns chunks of 1--64 actions.
\end{carditems}}
\end{minipage}\hfill\begin{minipage}[t]{0.49\linewidth}\vspace{0pt}
\cardpanel{cardgreen}{Evaluation rubric}{%
\begin{carditems}
\item Reward is success rate over the fixed episode set; missing episodes are failures.
\item Open tracks use official LIBERO-10 states or held-out clean RoboTwin scenes. Robust tracks use perturbations.
\item Failed replay, artifact validation, or completion thresholds yield zero.
\item No standard-performance gate for robustness tracks; no efficiency bonus.
\end{carditems}}
\cardpanel{cardrose}{Budget / evaluation protocol}{%
\begin{carditems}
\item \textbf{Develop:} 4 h, 1 H100, 48 CPUs, 64 GiB; up to 20,000 requested development episodes.
\item \textbf{Replay:} 1,800 s plus 120 s infrastructure allowance; training runs once offline.
\item LIBERO: 8-D state / 7-D action. RoboTwin: 16-D state / 14-D joint targets. Serving is separately time-limited.
\end{carditems}}
\end{minipage}\end{minipage}}\endgroup

%% file: content/task_cards/task06.tex
\begingroup
\hypersetup{hidelinks}
\setlength{\parindent}{0pt}\setlength{\parskip}{0pt}
\fontsize{9}{10.8}\selectfont
\setlength{\fboxsep}{9pt}\setlength{\fboxrule}{0.5pt}
\fcolorbox{cardline}{white}{%
\begin{minipage}{\dimexpr\linewidth-19pt\relax}
\cardpanel{cardhead}{}{%
\vspace{-9pt}{\large\bfseries Task Family 06: Blind Planar Pose Estimation}\hfill{\small\bfseries 4 instances}}
\textbf{Family:} Perception and estimation\hfill\textbf{Environment:} MuJoCo\par
\textbf{Task matrix:} RGB only, RGB-D, learned RGB-D, method agnostic\par\vspace{6pt}
\cardpanel{cardcream}{Agent Instructions (partial)}{%
A robot-mounted stick pushes one of three red, asymmetric blocks on a table. From a fixed oblique camera, \textbf{estimate the block\textquotesingle s planar pose and identify its shape}. Predictions should remain accurate through rotation and occlusion, with efficient CPU inference. [\ldots] Meshes and the private scene are not available; learn shape geometry and orientation conventions from rendered examples.}
\cardpanel{cardgray}{Environment / execution view}{%
\includegraphics[width=.325\linewidth]{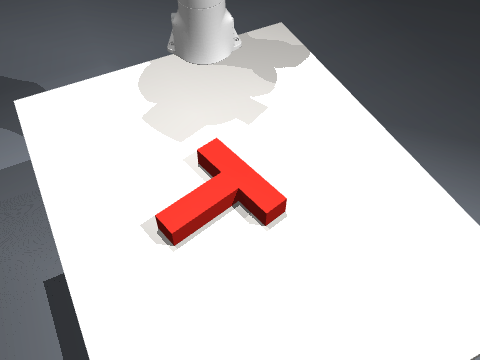}\hfill\includegraphics[width=.325\linewidth]{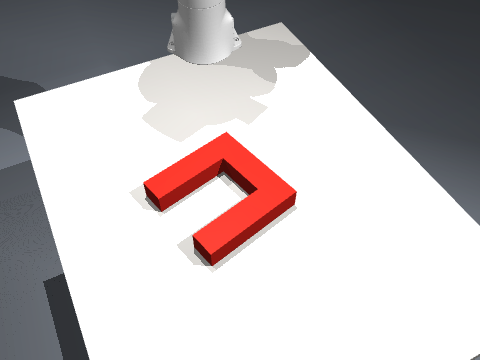}\hfill\includegraphics[width=.325\linewidth]{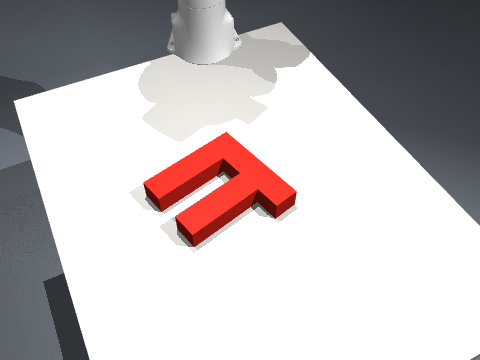}\par\vspace{3pt}\textbf{Task-scene renders:} T, C, and F blocks (left to right), viewed by the fixed oblique sensor camera. These geometry illustrations are not estimator predictions.\par\vspace{3pt}\begin{tabularx}{\linewidth}{@{}l@{\hspace{8pt}}X@{}}
A / B & RGB-only / RGB-D Python estimators; 4 CPUs.\\
C & RGB-D model trained from scratch; 16 CPUs, 1 GPU.\\
D & Method-agnostic Python estimator; 4 CPUs, 1 GPU.\\
\end{tabularx}}
\begin{minipage}[t]{0.49\linewidth}\vspace{0pt}
\cardpanel{cardblue}{Provided harness}{%
\begin{carditems}
\item Public deterministic renderers for individual frames and push episodes.
\item Noisy pose labels, camera calibration, time, and RGB; depth is variant-dependent. Meshes and shape labels are withheld.
\item Up to six design-set evaluation calls. All final inference runs on CPU.
\end{carditems}}
\cardpanel{cardblue}{Submitted artifacts}{%
\begin{carditems}
\item A / B / D: \texttt{estimator.py} with causal \texttt{reset} and \texttt{update}; return finite $(x,y,\theta,\mathrm{shape\_id})$.
\item C: CPU TorchScript \texttt{model.pt}, at most 20M tensor elements; independent per-frame predictions from RGB, depth, and validity mask.
\item Shape IDs 0, 1, 2 denote T, C, F.
\end{carditems}}
\end{minipage}\hfill\begin{minipage}[t]{0.49\linewidth}\vspace{0pt}
\cardpanel{cardgreen}{Evaluation rubric}{%
Each group uses the five largest translation and wrapped-angle errors, averaged separately. If all shape IDs are correct:\[g=\tfrac12 e^{-(\bar e_{xy}/0.01)^2}+\tfrac12 e^{-(\bar e_\theta/0.1)^2}.\]Otherwise $g=0$. Overall reward is\[R=\max(0,\,0.3\bar g_A+0.7\bar g_B-D).\]$D=0.2\operatorname{clip}_{[0,1]}[(10-\nu)/9]$, where $\nu$ is verifier-measured frames/s. A failed load or interface gate scores zero.}
\cardpanel{cardrose}{Budget / evaluation protocol}{%
\begin{carditems}
\item \textbf{Develop:} 2 h; resources vary by track as listed above.
\item \textbf{Evaluate:} 100 independent frames in ten groups, then ten push trajectories of at most 60 frames.
\item Python estimators may retain causal state. The learned model is stateless.
\end{carditems}}
\end{minipage}\end{minipage}}\endgroup

%% file: content/task_cards/task07.tex
\begingroup
\hypersetup{hidelinks}
\setlength{\parindent}{0pt}\setlength{\parskip}{0pt}
\fontsize{9}{10.8}\selectfont
\setlength{\fboxsep}{9pt}\setlength{\fboxrule}{0.5pt}
\fcolorbox{cardline}{white}{%
\begin{minipage}{\dimexpr\linewidth-19pt\relax}
\cardpanel{cardhead}{}{%
\vspace{-9pt}{\large\bfseries Task Family 07: Contact-Rich Bin Clearing}\hfill{\small\bfseries 1 instance}}
\textbf{Family:} Perception and estimation\hfill\textbf{Environment:} MuJoCo\par
\textbf{Task matrix:} One policy evaluated on eight hidden piles\par\vspace{6pt}
\cardpanel{cardcream}{Agent Instructions (partial)}{%
A green KLT bin arrives at your robot cell holding 12--16 stamped steel U-brackets, dumped loose. A conveyor runs along the other side of the cell. Your job: ship the software that empties the bin onto the conveyor---\textbf{as many parts as possible, as fast as possible, breaking nothing}. [\ldots] The drop zone is a small placement nest [\ldots] and it takes ONE part at a time.}
\cardpanel{cardgray}{Environment / execution view}{%
{\centering\includegraphics[width=\linewidth,height=1.48in,keepaspectratio]{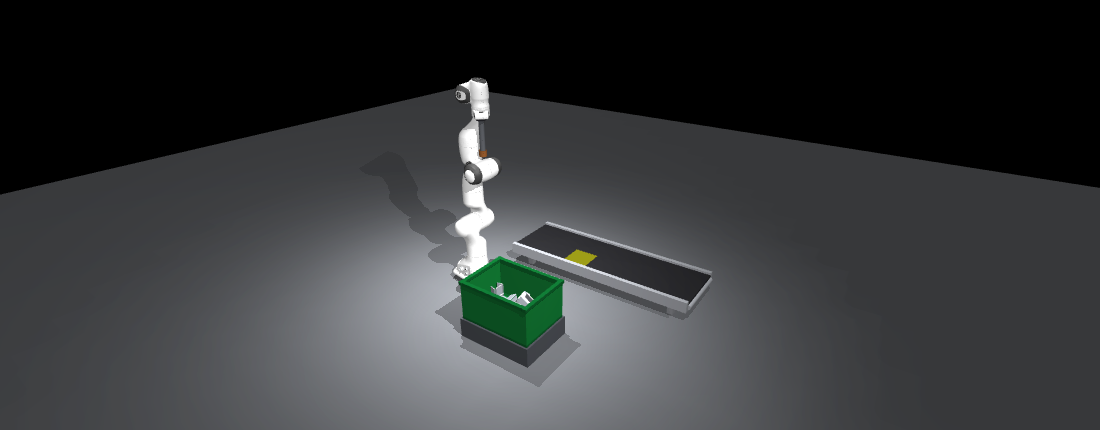}\par}\vspace{3pt}\textbf{Task-scene render:} A public pile (seed 101), contact electromagnet, and conveyor drop zone. Contact is required to pick up a bracket; delivering several together blocks the nest until extras are removed.}
\begin{minipage}[t]{0.49\linewidth}\vspace{0pt}
\cardpanel{cardblue}{Provided harness}{%
\begin{carditems}
\item Panda workcell, a contact electromagnet, public geometry and sensor calibration, and five public pile seeds.
\item 20 Hz joint positions, velocities, torques, wrist force/torque, magnet state, and time.
\item 10 Hz overhead and wrist RGB-D images. Ground-truth part poses are withheld.
\item A development runner, diagnostic events, and optional videos.
\end{carditems}}
\cardpanel{cardblue}{Submitted artifacts}{%
\begin{carditems}
\item An isolated \texttt{policy/} package exposing \texttt{make\_policy()}, \texttt{reset()}, and \texttt{act()}.
\item At each control tick: seven joint-position targets and one magnet command.
\item Clear brackets into the conveyor nest one at a time while avoiding floor drops, damage, and hard bin impacts.
\end{carditems}}
\end{minipage}\hfill\begin{minipage}[t]{0.49\linewidth}\vspace{0pt}
\cardpanel{cardgreen}{Evaluation rubric}{%
For cleared fraction $f$, throughput $p$, and clean complete-clear indicator $I$:\[\begin{aligned}s={}&0.30I+0.35h(f)\\ &+0.20\min(p/10,1)\\ &+0.15I\min(p/15,1)\\ &-0.03N_{\rm floor}-0.05N_{\rm damage}\\ &-0.05N_{\rm bin}.\end{aligned}\]$h(f)=f/6$ for $f\leq0.6$; otherwise $h(f)=0.1+0.9[(f-0.6)/0.4]^3$. Clip episode scores to $[0,1]$ and average over eight piles. Missing/unloadable policies score zero; subsequent smoke failure caps reward at 0.1.}
\cardpanel{cardrose}{Budget / evaluation protocol}{%
\begin{carditems}
\item \textbf{Develop:} 4 h, 4 CPUs.
\item \textbf{Evaluate:} Eight hidden piles, each with 120 s of simulated time and 360 s cumulative policy computation.
\item CPU-only inference; each call has a 30 s hang cap. Throughput uses clear makespan, or 120 s if incomplete.
\end{carditems}}
\end{minipage}\end{minipage}}\endgroup

%% file: content/task_cards/task08.tex
\begingroup
\hypersetup{hidelinks}
\setlength{\parindent}{0pt}\setlength{\parskip}{0pt}
\fontsize{9}{10.8}\selectfont
\setlength{\fboxsep}{9pt}\setlength{\fboxrule}{0.5pt}
\fcolorbox{cardline}{white}{%
\begin{minipage}{\dimexpr\linewidth-19pt\relax}
\cardpanel{cardhead}{}{%
\vspace{-9pt}{\large\bfseries Task Family 08: Universal Mobile-Manipulator Base}\hfill{\small\bfseries 1 instance}}
\textbf{Family:} Mechanical design\hfill\textbf{Environment:} MuJoCo\par
\textbf{Task matrix:} One chassis $\times$ three arms $\times$ 12 targets $\times$ two loads\par\vspace{6pt}
\cardpanel{cardcream}{Agent Instructions (partial)}{%
Design and build one mecanum mobile-manipulator base compatible with the three supplied canonical arms: Franka Panda, Universal Robots UR5e, and UFACTORY xArm7. \textbf{The same submitted chassis, battery, wheels, and \texttt{arm\_mount\_site} must be used unchanged with every arm.} [\ldots] Submit \texttt{controller.py} to control shelf entry and loaded target holding.}
\cardpanel{cardgray}{Environment / execution view}{%
\begin{minipage}{.52\linewidth}\raggedright {\centering\includegraphics[width=\linewidth,height=1.6in,keepaspectratio]{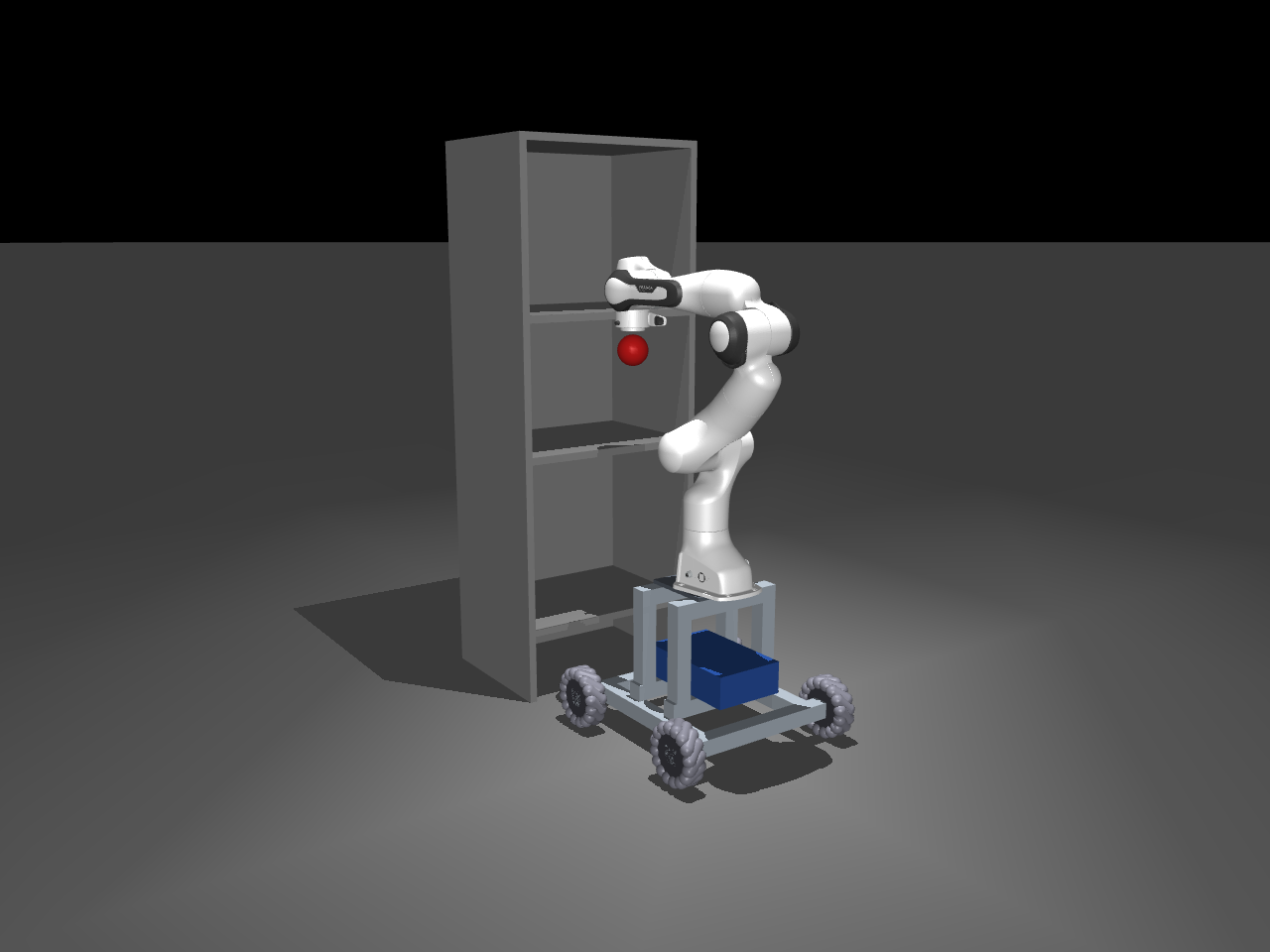}\par}\end{minipage}\hfill\begin{minipage}{.45\linewidth}\raggedright \textbf{Example assembly.} Existing project visualization of a mobile base, Panda arm, payload, and shelf.\par\vspace{5pt}The verifier replaces the preview arm with each canonical model. The base and mounting interface remain unchanged.\par\vspace{5pt}\textbf{Required load:} 1 kg.\par\textbf{Margin test:} 2 kg.\par All 12 shelf targets count for each arm.\end{minipage}}
\begin{minipage}[t]{0.49\linewidth}\vspace{0pt}
\cardpanel{cardblue}{Provided harness}{%
\begin{carditems}
\item Canonical Panda (7-DoF), UR5e (6-DoF), and xArm7 (7-DoF) models.
\item A fixed universal adapter, immutable mecanum wheels, battery, payload, aluminum-profile catalog, and shelf specification.
\item A preview scene for testing the submitted base and controller.
\end{carditems}}
\cardpanel{cardblue}{Submitted artifacts}{%
\begin{carditems}
\item \texttt{robot.xml} and \texttt{controller.py}, with required model names and mounting site.
\item At 50 Hz, the controller maps time, joint state, and current controls to one finite command per actuator.
\item The submitted controller performs shelf entry and holding; verifier controllers run other design tests.
\end{carditems}}
\end{minipage}\hfill\begin{minipage}[t]{0.49\linewidth}\vspace{0pt}
\cardpanel{cardgreen}{Evaluation rubric}{%
Take the \textbf{minimum across the three arms at each checkpoint}, then apply weights:\par\vspace{3pt}\begin{tabularx}{\linewidth}{@{}l@{\hspace{8pt}}X@{}}
Validity & 15\%\\
Design / resources & 35\%\\
Static stability & 15\%\\
Dynamic stability & 20\%\\
Shelf integration & 15\%\\
\end{tabularx}\par\vspace{3pt}Shelf integration assigns 10\% to 1 kg and 5\% to 2 kg trials. A failed load or structure checkpoint caps reward at 0.15.}
\cardpanel{cardrose}{Budget / evaluation protocol}{%
\begin{carditems}
\item \textbf{Develop:} 2 h, 4 CPUs.
\item Three trusted arm assemblies; all 12 targets at both loads, plus static and dynamic batteries.
\item Compactness, mass, and material use affect design credit. Reach and hold without shelf collision, tipping, or sustained loss of wheel support.
\end{carditems}}
\end{minipage}\end{minipage}}\endgroup

%% file: content/task_cards/task09.tex
\begingroup
\hypersetup{hidelinks}
\setlength{\parindent}{0pt}\setlength{\parskip}{0pt}
\fontsize{9}{10.8}\selectfont
\setlength{\fboxsep}{9pt}\setlength{\fboxrule}{0.5pt}
\fcolorbox{cardline}{white}{%
\begin{minipage}{\dimexpr\linewidth-19pt\relax}
\cardpanel{cardhead}{}{%
\vspace{-9pt}{\large\bfseries Task Family 09: Multi-Robot GELLO Co-Design}\hfill{\small\bfseries 1 instance}}
\textbf{Family:} Mechanical design\hfill\textbf{Environment:} MuJoCo\par
\textbf{Task matrix:} Three robot-matched leader devices, developed together\par\vspace{6pt}
\cardpanel{cardcream}{Agent Instructions (partial)}{%
Design gravity compensation for three CAD-assembled GELLO lead arms, one for each of these follower robots: Franka Panda, Universal Robots UR5e, and UFACTORY xArm7. [\ldots] Modify \texttt{lead.xml} with springs, masses, supports, and counterweights so the passive arm holds pose while remaining easy to move. [\ldots] Implement \texttt{trim.py} with model-based feedforward and per-device adaptation from probe measurements.}
\cardpanel{cardgray}{Environment / execution view}{%
\begin{minipage}{.6\linewidth}\raggedright {\centering\includegraphics[width=\linewidth,height=1.65in,keepaspectratio,trim=480bp 70bp 480bp 114bp,clip]{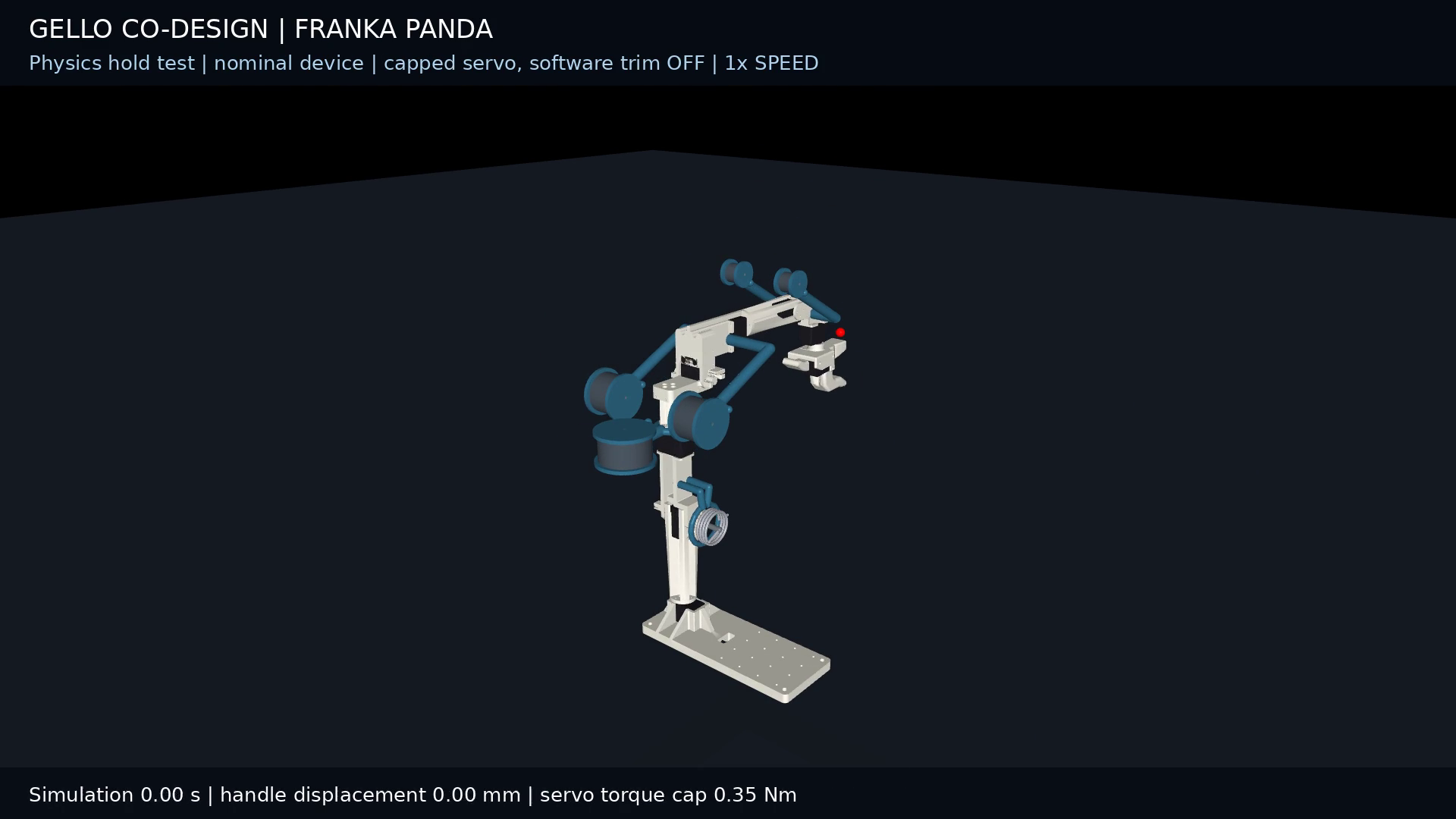}\par}\end{minipage}\hfill\begin{minipage}{.37\linewidth}\raggedright \textbf{Example: Franka GELLO.} A submitted gravity-compensation mechanism in simulation. The frame shows the submitted leader arm and counterweights.\par\vspace{5pt}\textbf{Variants:} Franka (7 joints), UR5e (6), and xArm7 (7).\end{minipage}}
\begin{minipage}[t]{0.49\linewidth}\vspace{0pt}
\cardpanel{cardblue}{Provided harness}{%
\begin{carditems}
\item Measured geometry, uncompensated leader assemblies, workspace grids, nominal dynamics utilities, and public analysis helpers.
\item Preserve joint frames, geometry, mounts, and stock servos; obey mass and printability constraints.
\item Hidden device copies vary link masses and handle payload.
\end{carditems}}
\cardpanel{cardblue}{Submitted artifacts}{%
\begin{carditems}
\item For each robot: modified \texttt{lead.xml}, \texttt{trim.py}, and referenced meshes.
\item Passive compensation plus nominal and adapted feedforward torque functions.
\item \texttt{plan\_probe} selects 15 additional poses after a home measurement; \texttt{adapt} receives all 16 settled-pose probes.
\end{carditems}}
\end{minipage}\hfill\begin{minipage}[t]{0.49\linewidth}\vspace{0pt}
\cardpanel{cardgreen}{Evaluation rubric}{%
Each variant has ten equally weighted reward checkpoints:\par\vspace{3pt}\begin{carditems}
\item Motion clearance: 10\%.
\item Passive residual torque, hold, and backdrive: 30\%.
\item Adapted reference-arm hold and backdrive: 20\%.
\item Combined hold, backdrive, servo headroom, and push recovery: 40\%.
\end{carditems}\vspace{3pt}Validity shortfalls deduct up to 0.10; floor at zero. Load/structure failure skips the submitted-arm battery and unavailable metrics score zero. Final reward averages all three variants.}
\cardpanel{cardrose}{Budget / evaluation protocol}{%
\begin{carditems}
\item \textbf{Develop:} 3 h, 4 CPUs for all three designs.
\item Maximum device mass: 2.30 kg; stock servo torque limit: 0.35 Nm.
\item Probe poses are selected as one batch. The verifier freezes sampled feedforward during each scenario and tests submitted and reference assemblies.
\end{carditems}}
\end{minipage}\end{minipage}}\endgroup

%% file: content/appendix_full_results.tex
\clearpage
\section{Full Per-Task Results}
\label{app:fullresults}
\input{content/generated/appendix_counts}

\subsection{Coverage and Reading Guide}
\label{app:results_coverage}

We report all 51 tasks across nine task families and four workflows for the
\appendixSystems{} model--harness systems.
All scores below are verifier rewards multiplied by 100, not uniformly success
percentages. Family scores average their constituent tasks. The RLE Index first
averages family scores within each workflow (01--03, 04--05, 06--07, and 08--09),
then averages the four workflow scores. Every plot uses the same descending
RLE Index order. Numeric annotations are rounded to one decimal; the archived
export and accompanying CSV preserve the exported precision and all native
metrics. Missing values are marked NA or a dash, distinct from measured zeros.

\begin{figure}[H]
\centering
\includegraphics[width=\linewidth]{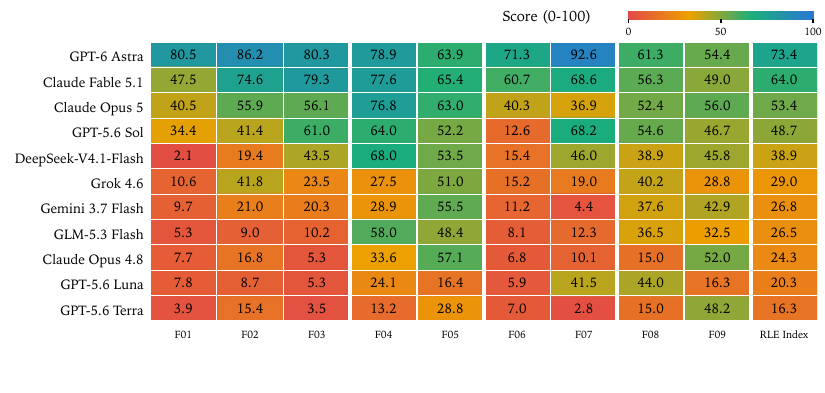}
\caption{Task-family scores (F01--F09) and overall RLE Index for every system.
The same 0--100 scale is used throughout the reward figures. The Index gives
workflows equal weight, so it is not the unweighted mean of these nine columns.}
\label{fig:appendix_overview}
\end{figure}

\clearpage
\label{app:results_kitchen}

\begin{figure}[h]
\centering
\includegraphics[width=\linewidth]{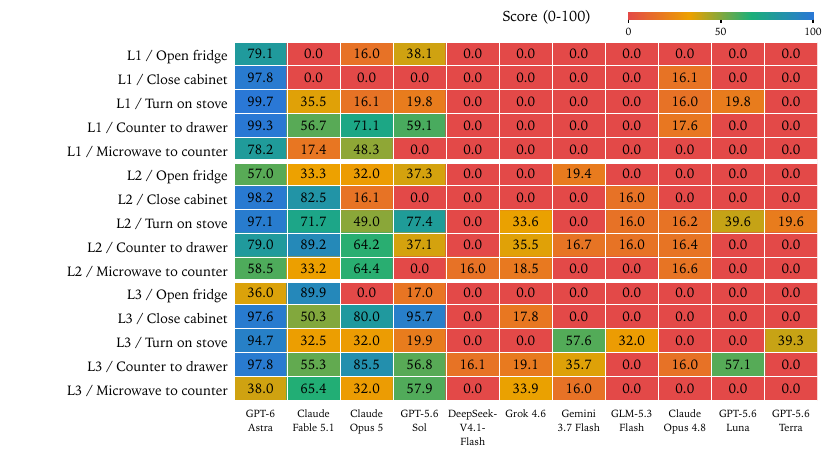}
\caption{\textbf{Family 01: agentic control.} All 15 tasks: five kitchen activities
at each of three scaffolding levels. L1 provides the action API, L2 adds the
harness library, and L3 adds privileged state. Each cell is one selected run's
verifier reward; horizontal separators distinguish levels.}
\label{fig:appendix_family01}
\end{figure}

\begin{figure}[H]
\centering
\includegraphics[width=\linewidth]{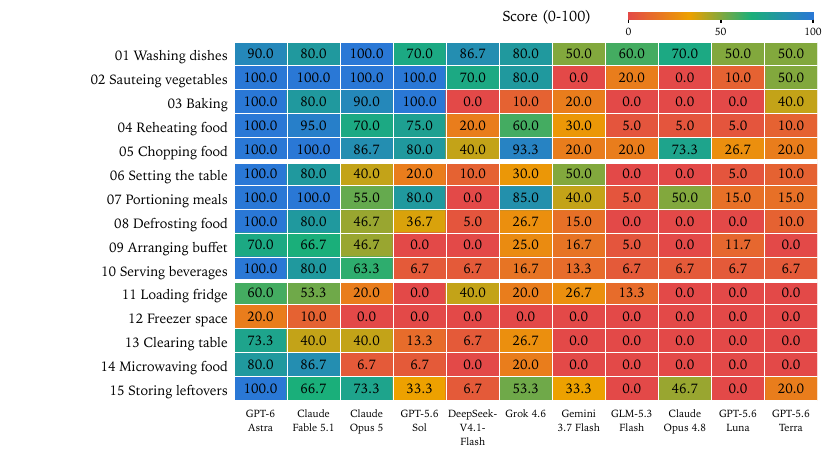}
\caption{\textbf{Family 02: harness transfer.} All 15 activity groups, with
separators after the five easy and five medium groups; the final five are hard.
Each reward measures stage credit achieved by fresh agents on the group's
held-out task using the submitted harness. Group numbers match the task catalog.}
\label{fig:appendix_family02}
\end{figure}

\label{app:results_physical}

\begin{figure}[H]
\centering
\includegraphics[width=\linewidth]{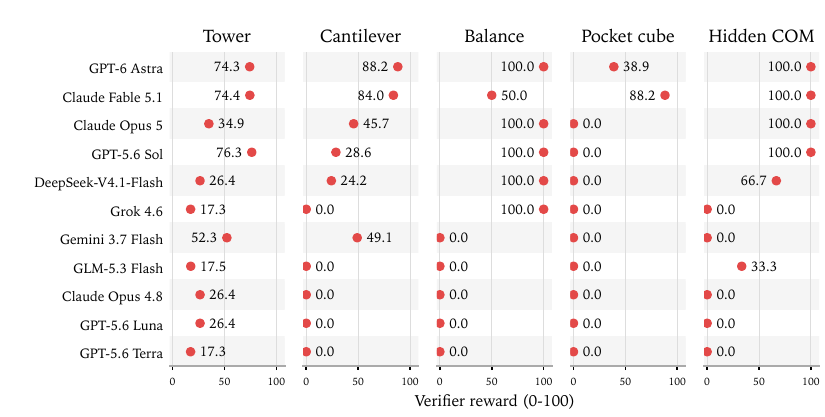}
\caption{\textbf{Family 03: tabletop physical reasoning.} Dot plots show all five
task rewards separately, with numeric annotations. The balance task uses the
task identifier \texttt{03-balance-coins}; its audited task card specifies
heavy-cube identification. Hidden COM denotes hidden center of mass.
Values use the task-specific verifier reward, not a common physical unit.}
\label{fig:appendix_family03}
\end{figure}

\begin{figure}[H]
\centering
\includegraphics[width=\linewidth]{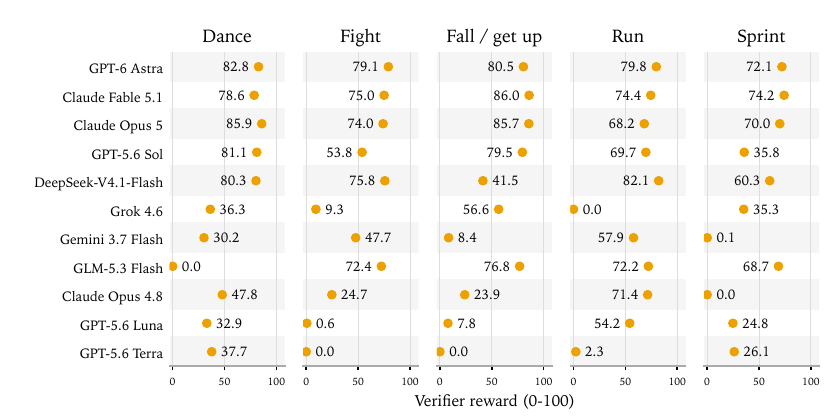}
\caption{\textbf{Family 04: whole-body motion tracking.} Rewards for all five
motion clips. Each dot represents the selected run's final verifier reward,
which combines tracking and survival. Shared axes expose differences across
clips without averaging them away. Native MPJPE and survival are reported in
Table~\ref{tab:appendix_native}.}
\label{fig:appendix_family04}
\end{figure}

The two families retain distinct evaluation contracts. Family~03 scores physical
outcomes or submitted answers within an interaction attempt; Family~04 evaluates
a submitted tracking policy on hidden seeds. Dot plots place each task on its
own axis while retaining the common normalized reward scale.

\clearpage
\label{app:results_policy_pose}

\begin{figure}[H]
\centering
\includegraphics[width=\linewidth]{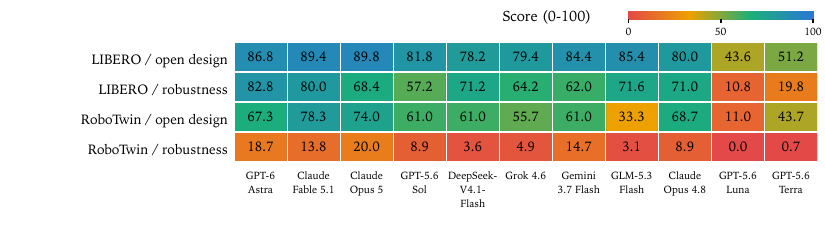}
\caption{\textbf{Family 05: nanoVLA recipe engineering.} All four LIBERO and
RoboTwin tracks, separating open design from robustness. Entries are final
hidden-evaluation rewards, not the best development-time evaluation scores.}
\label{fig:appendix_family05}
\end{figure}

\begin{figure}[H]
\centering
\includegraphics[width=\linewidth]{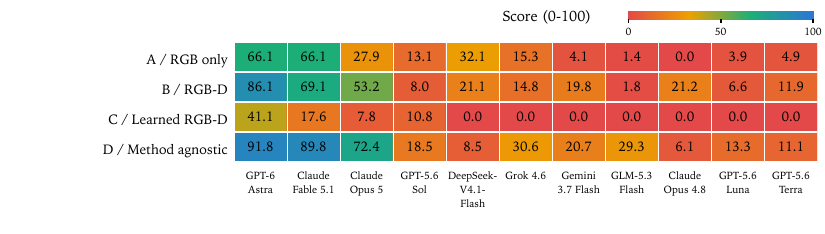}
\caption{\textbf{Family 06: blind planar pose estimation.} All four estimator
variants. A--D match the task card: RGB-only, RGB-D, learned RGB-D, and
method-agnostic estimation. Entries are final rewards after the verifier
inference-speed deduction and load/interface gate; they are not raw pose accuracy.}
\label{fig:appendix_family06}
\end{figure}

\input{content/generated/appendix_native}

\paragraph{Interpreting native diagnostics.}
Table~\ref{tab:appendix_native} supplements normalized rewards with physical
units. Motion errors and survival are averaged over clips, whereas bin-clearance
and throughput are copied from the joint task report. These summaries do not
reconstruct gated rewards. The machine-readable records retain every exported
per-task metric, including pose errors, inference rates, gate fields, and
mechanical-design checkpoints.

\clearpage

\begin{figure}[H]
\centering
\includegraphics[width=\linewidth]{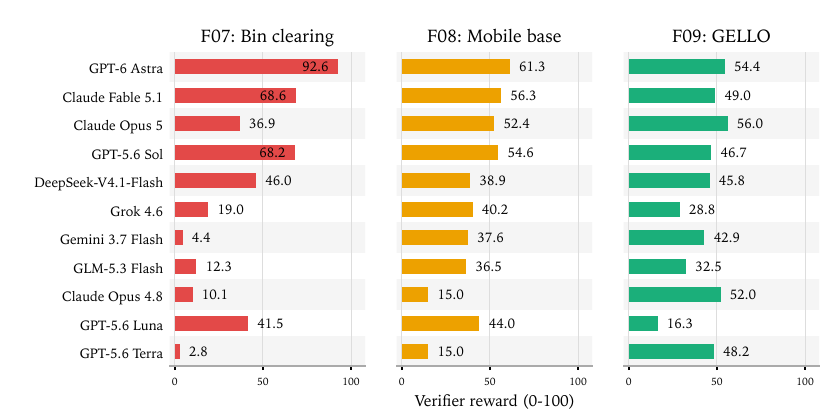}
\caption{\textbf{Families 07--09: complete joint-task rewards.} Horizontal bars
show bin clearing, mobile-base design, and GELLO gravity compensation.
Each family contains one submitted joint task per system. Family~08 uses
checkpoint-wise worst-arm credit and a gate cap; Family~09 averages the three
arm rewards. Bars show the reported final reward with all gates applied.}
\label{fig:appendix_families07_09}
\end{figure}

\begin{figure}[H]
\centering
\includegraphics[width=\linewidth]{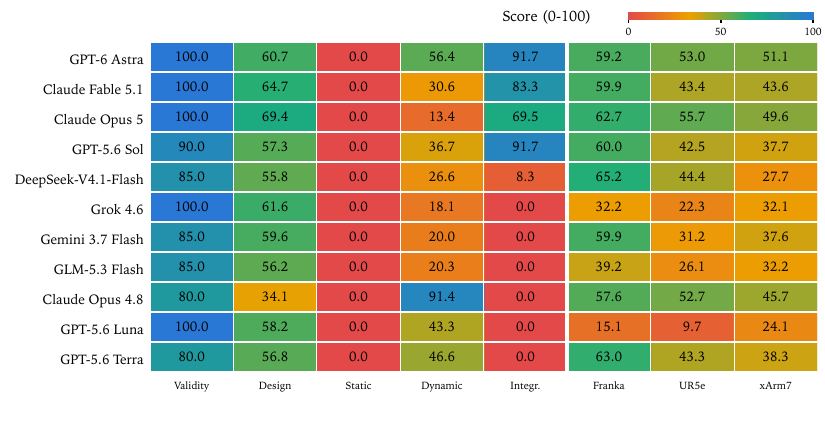}
\caption{\textbf{Mechanical-design diagnostics.} The first five columns are
Family~08 stage credits divided by their stage weights (0.15, 0.35, 0.15,
0.20, 0.15) and multiplied by 100. They precede the final gate cap and must
not be substituted for the reward above. The last three columns are
Family~09's individual arm rewards (0--100), whose arithmetic mean is the
joint-task reward. Integr. denotes integration.}
\label{fig:appendix_design_diagnostics}
\end{figure}

\begin{figure}[H]
\centering
\includegraphics[width=\linewidth]{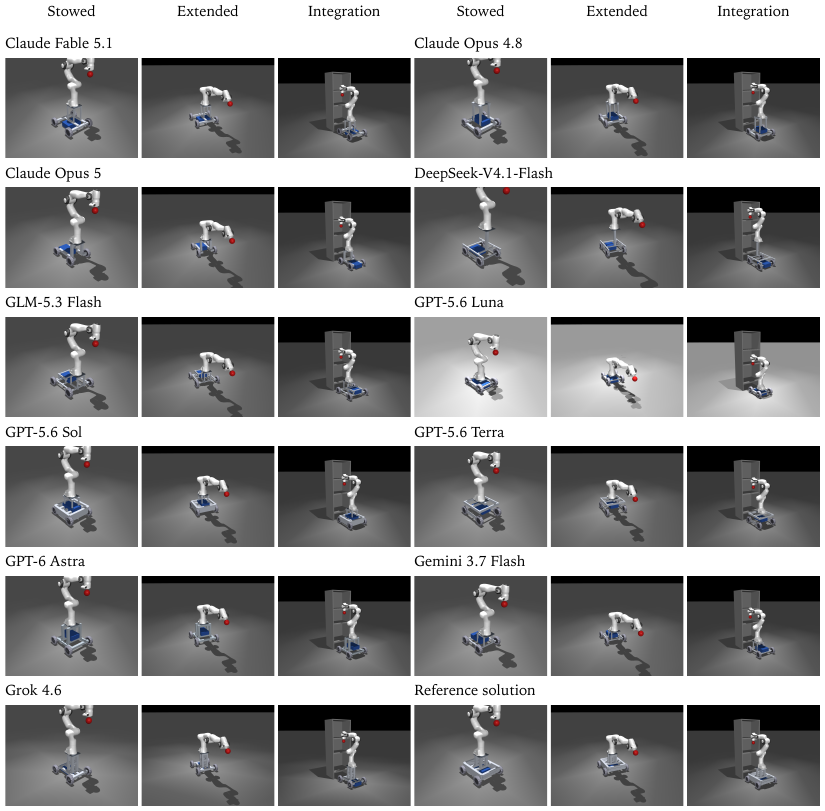}
\caption{\textbf{Family 08: mobile-base design renderings.} The 36 panels show
designs from all eleven evaluated models and a reference solution. Each labeled
group contains stowed, extended, and integration views of one design.
Verifier camera settings are preserved.}
\label{fig:appendix_task08_designs}
\end{figure}

\clearpage
\paragraph{Design renderings.}
Figure~\ref{fig:appendix_task08_designs} shows mobile-base designs, organized
by model, alongside a reference solution.

\clearpage
\subsection{Resource Use}
\label{app:results_resources}

\input{content/generated/appendix_resources}

\paragraph{Cost and elapsed time.}
Mean API cost and elapsed time follow the same task--family--workflow hierarchy
as the RLE Index. Total cost instead sums all retained runs. 
Elapsed time is the difference between run start and finish timestamps, not
isolated agent compute. Failed-status
counts describe run execution status, where all of them are timeout failures where the agent does not call termination or submission until the whole time budget is used up.

%% file: content/generated/appendix_counts.tex
\newcommand{\appendixSystems}{11}
\newcommand{\appendixRuns}{561}
\newcommand{\appendixCompleted}{552}
\newcommand{\appendixFailed}{9}

%% file: content/generated/appendix_native.tex
\begin{table}[H]
\centering\footnotesize
\setlength{\tabcolsep}{3pt}
\caption{Native diagnostics. MPJPE (lower is better) and survival are arithmetic means over the five motion clips in Family~04. Clearance and throughput are reported for Family~07. Higher is better for the latter three columns; these quantities are not interchangeable with verifier reward.}
\label{tab:appendix_native}
\begin{tabular}{@{}lrrrr@{}}
\toprule
System & MPJPE (mm) & Survival (\%) & Clearance (\%) & Parts/min \\
\midrule
GPT-6 Astra & 32.62 & 98.71 & 100.00 & 9.22 \\
Claude Fable 5.1 & 40.42 & 100.00 & 86.90 & 14.58 \\
Claude Opus 5 & 34.90 & 93.39 & 78.69 & 6.02 \\
GPT-5.6 Sol & 51.50 & 87.03 & 96.69 & 8.35 \\
DeepSeek-V4.1-Flash & 43.52 & 91.41 & 87.09 & 6.82 \\
Grok 4.6 & 62.22 & 44.77 & 71.52 & 5.89 \\
Gemini 3.7 Flash & 62.48 & 46.41 & 27.32 & 10.62 \\
GLM-5.3 Flash & 55.64 & 77.94 & 61.17 & 6.02 \\
Claude Opus 4.8 & 69.82 & 51.27 & 52.74 & 3.92 \\
GPT-5.6 Luna & 79.44 & 41.43 & 90.70 & 6.80 \\
GPT-5.6 Terra & 89.38 & 26.54 & 29.33 & 4.08 \\
\bottomrule
\end{tabular}
\end{table}

%% file: content/generated/appendix_resources.tex
\begin{table}[H]
\centering\footnotesize
\setlength{\tabcolsep}{3pt}
\caption{Resources and status for the retained runs. Costs are USD; elapsed time is in hours; token totals are in millions. Cost coverage counts runs with reported API costs. Grok 4.6 and Gemini 3.7 Flash token counts and total cost are absent due to the harness incompatibility with Harbor.}
\label{tab:appendix_resources}
\begin{tabular}{@{}lrrrrrrrr@{}}
\toprule
System & Mean \$ & Total \$ & Coverage & Mean h & Input M & Cached M & Output M & Failed \\
\midrule
GPT-6 Astra & 39.00 & 3899.09 & 51/51 & 2.42 & 2601.41 & 2552.76 & 12.07 & 0 \\
Claude Fable 5.1 & 22.81 & 2246.63 & 51/51 & 2.50 & 2671.29 & 2630.17 & 22.51 & 0 \\
Claude Opus 5 & 27.01 & 3143.92 & 51/51 & 2.53 & 5026.40 & 4968.04 & 27.91 & 0 \\
GPT-5.6 Sol & 17.17 & 1535.96 & 51/51 & 2.26 & 2872.33 & 2821.53 & 8.59 & 0 \\
DeepSeek-V4.1-Flash & 0.40 & 39.34 & 51/51 & 2.31 & 3372.05 & 3357.09 & 23.19 & 0 \\
GLM-5.3 Flash & 1.16 & 118.52 & 51/51 & 4.93 & 2994.11 & 2891.37 & 32.74 & 7 \\
Claude Opus 4.8 & 22.79 & 2034.56 & 51/51 & 2.33 & 3105.33 & 3066.26 & 25.53 & 0 \\
GPT-5.6 Luna & 0.52 & 43.07 & 51/51 & 1.44 & 1525.54 & 1496.75 & 6.14 & 0 \\
GPT-5.6 Terra & 2.35 & 257.79 & 51/51 & 1.06 & 837.97 & 816.77 & 4.33 & 1 \\
\bottomrule
\end{tabular}
\end{table}

%% file: main.bbl
\begin{thebibliography}{43}
\providecommand{\natexlab}[1]{#1}
\providecommand{\url}[1]{\texttt{#1}}
\expandafter\ifx\csname urlstyle\endcsname\relax
  \providecommand{\doi}[1]{doi: #1}\else
  \providecommand{\doi}{doi: \begingroup \urlstyle{rm}\Url}\fi

\bibitem[{Anthropic}(2026{\natexlab{a}})]{anthropic2026fable51}
{Anthropic}.
\newblock {Claude Fable 5.1} system card, September 2026{\natexlab{a}}.
\newblock URL
  \url{https://www-cdn.anthropic.com/0339e6a7c5c7b87f5c07798616dc32c215d14235/Claude%20Fable%205.1%20&%20Claude%20Mythos%205.1%20System%20Card.pdf}.

\bibitem[{Anthropic}(2026{\natexlab{b}})]{anthropic2026opus48}
{Anthropic}.
\newblock {Claude Opus 4.8} system card, May 2026{\natexlab{b}}.
\newblock URL \url{https://anthropic.com/claude-opus-4-8-system-card}.

\bibitem[{Anthropic}(2026{\natexlab{c}})]{anthropic2026opus5}
{Anthropic}.
\newblock {Claude Opus 5} system card, July 2026{\natexlab{c}}.
\newblock URL \url{https://anthropic.com/claude-opus-5-system-card}.

\bibitem[Black et~al.(2024)Black, Brown, Driess, Esmail, Equi, Finn, Fusai,
  Groom, Hausman, Ichter, et~al.]{black2024pi_0}
Kevin Black, Noah Brown, Danny Driess, Adnan Esmail, Michael Equi, Chelsea
  Finn, Niccolo Fusai, Lachy Groom, Karol Hausman, Brian Ichter, et~al.
\newblock {$\pi_0$}: A vision-language-action flow model for general robot
  control.
\newblock \emph{arXiv preprint arXiv:2410.24164}, 2024.

\bibitem[Brockman et~al.(2016)Brockman, Cheung, Pettersson, Schneider,
  Schulman, Tang, and Zaremba]{brockman2016openai}
Greg Brockman, Vicki Cheung, Ludwig Pettersson, Jonas Schneider, John Schulman,
  Jie Tang, and Wojciech Zaremba.
\newblock Openai gym.
\newblock \emph{arXiv preprint arXiv:1606.01540}, 2016.

\bibitem[Brohan et~al.(2022)Brohan, Brown, Carbajal, Chebotar, Dabis, Finn,
  Gopalakrishnan, Hausman, Herzog, Hsu, et~al.]{brohan2022rt}
Anthony Brohan, Noah Brown, Justice Carbajal, Yevgen Chebotar, Joseph Dabis,
  Chelsea Finn, Keerthana Gopalakrishnan, Karol Hausman, Alex Herzog, Jasmine
  Hsu, et~al.
\newblock Rt-1: Robotics transformer for real-world control at scale.
\newblock \emph{arXiv preprint arXiv:2212.06817}, 2022.

\bibitem[Carion et~al.(2026)Carion, Gustafson, Hu, Debnath, Hu, Suris
  Coll-Vinent, Ryali, Alwala, Khedr, Huang, et~al.]{carion2026sam}
Nicolas Carion, Laura Gustafson, Yuan-Ting Hu, Shoubhik Debnath, Ronghang Hu,
  Didac Suris Coll-Vinent, Chaitanya Ryali, Kalyan~Vasudev Alwala, Haitham
  Khedr, Andrew Huang, et~al.
\newblock Sam 3: Segment anything with concepts.
\newblock In \emph{International conference on learning representations},
  volume 2026, pp.\  138846--138923, 2026.

\bibitem[Chan et~al.(2025)Chan, Chowdhury, Jaffe, Aung, Sherburn, Mays,
  Starace, Liu, Maksin, Patwardhan, et~al.]{chan2025mle}
Jun~Shern Chan, Neil Chowdhury, Oliver Jaffe, James Aung, Dane Sherburn, Evan
  Mays, Giulio Starace, Kevin Liu, Leon Maksin, Tejal Patwardhan, et~al.
\newblock Mle-bench: Evaluating machine learning agents on machine learning
  engineering.
\newblock In \emph{International Conference on Learning Representations},
  volume 2025, pp.\  50466--50494, 2025.

\bibitem[Chen et~al.(2025)Chen, Chen, Chen, Cai, Liu, Li, Liang, Lin, Ge, Gu,
  et~al.]{chen2025robotwin}
Tianxing Chen, Zanxin Chen, Baijun Chen, Zijian Cai, Yibin Liu, Zixuan Li,
  Qiwei Liang, Xianliang Lin, Yiheng Ge, Zhenyu Gu, et~al.
\newblock Robotwin 2.0: A scalable data generator and benchmark with strong
  domain randomization for robust bimanual robotic manipulation.
\newblock \emph{arXiv preprint arXiv:2506.18088}, 2025.

\bibitem[Chen et~al.(2026)Chen, Chen, Li, Tang, Su, Lu, Wan, Chen, Liu, Yan,
  et~al.]{chen2026robodojo}
Tianxing Chen, Yue Chen, Zixuan Li, Junyuan Tang, Kailun Su, Haoran Lu, Weijie
  Wan, Baijun Chen, Songling Liu, Haowen Yan, et~al.
\newblock Robodojo: A unified sim-and-real benchmark for comprehensive
  evaluation of generalist robot manipulation policies.
\newblock \emph{arXiv preprint arXiv:2607.04434}, 2026.

\bibitem[{DeepSeek-AI}(2026)]{deepseek2026v41flashcard}
{DeepSeek-AI}.
\newblock {DeepSeek-V4.1-Flash}.
\newblock Hugging Face model card, 2026.
\newblock URL \url{https://huggingface.co/deepseek-ai/DeepSeek-V4.1-Flash}.

\bibitem[Elmaaroufi et~al.(2026)Elmaaroufi, Svegliato, Kalade, Schelle, Seshia,
  and Zaharia]{elmaaroufi2026rho}
Karim Elmaaroufi, Justin Svegliato, Sarunas Kalade, Graham Schelle, Sanjit~A
  Seshia, and Matei Zaharia.
\newblock Rho: Your coding agent is secretly a roboticist.
\newblock \emph{arXiv preprint arXiv:2606.16458}, 2026.

\bibitem[Fu et~al.(2026)Fu, Yu, El-Refai, Kou, Xue, Huang, Xiao, Wang, Niu, Li,
  et~al.]{fu2026capx}
Letian Fu, Justin Yu, Karim El-Refai, Ethan Kou, Haoru Xue, Huang Huang, Wenli
  Xiao, Guanzhi Wang, Dantong Niu, Fei-Fei Li, et~al.
\newblock Cap-x: A framework for benchmarking and improving coding agents for
  robot manipulation.
\newblock \emph{arXiv preprint arXiv:2603.22435}, 2026.

\bibitem[{Google DeepMind}(2026)]{google2026gemini37flash}
{Google DeepMind}.
\newblock {Gemini 3.7 Flash} model card, August 2026.
\newblock URL
  \url{https://deepmind.google/models/model-cards/gemini-3-7-flash/}.

\bibitem[{Grok}(2026)]{grok46}
{Grok}.
\newblock {Grok 4.6} system card, August 2026.
\newblock URL \url{https://media.x.ai/v1/website/card-4p6-4cd2dc57.pdf}.

\bibitem[Harvey et~al.(2020)Harvey, Yurick, Nowrouzezahrai, and
  Pal]{harvey2020robust}
Félix~G. Harvey, Mike Yurick, Derek Nowrouzezahrai, and Christopher Pal.
\newblock Robust motion in-betweening.
\newblock \emph{ACM Transactions on Graphics (Proceedings of ACM SIGGRAPH)},
  39\penalty0 (4), 2020.

\bibitem[Huang et~al.(2023{\natexlab{a}})Huang, Vora, Liang, and
  Leskovec]{huang2023mlagentbench}
Qian Huang, Jian Vora, Percy Liang, and Jure Leskovec.
\newblock Mlagentbench: Evaluating language agents on machine learning
  experimentation.
\newblock \emph{arXiv preprint arXiv:2310.03302}, 2023{\natexlab{a}}.

\bibitem[Huang et~al.(2023{\natexlab{b}})Huang, Wang, Zhang, Li, Wu, and
  Fei-Fei]{huang2023voxposer}
Wenlong Huang, Chen Wang, Ruohan Zhang, Yunzhu Li, Jiajun Wu, and Li~Fei-Fei.
\newblock Voxposer: Composable 3d value maps for robotic manipulation with
  language models.
\newblock \emph{arXiv preprint arXiv:2307.05973}, 2023{\natexlab{b}}.

\bibitem[James et~al.(2020)James, Ma, Arrojo, and Davison]{james2020rlbench}
Stephen James, Zicong Ma, David~Rovick Arrojo, and Andrew~J Davison.
\newblock Rlbench: The robot learning benchmark \& learning environment.
\newblock \emph{IEEE Robotics and Automation Letters}, 5\penalty0 (2):\penalty0
  3019--3026, 2020.

\bibitem[Jimenez et~al.(2024)Jimenez, Yang, Wettig, Yao, Pei, Press, and
  Narasimhan]{jimenez2024swe}
Carlos~E Jimenez, John Yang, Alexander Wettig, Shunyu Yao, Kexin Pei, Ofir
  Press, and Karthik Narasimhan.
\newblock Swe-bench: Can language models resolve real-world github issues?
\newblock In \emph{International Conference on Learning Representations},
  volume 2024, pp.\  54107--54157, 2024.

\bibitem[Jin et~al.(2026)Jin, Guo, Jia, Deng, Li, Liu, Liao, Prasad, Franzius,
  Neumann, et~al.]{jin2026nautilus}
Yufeng Jin, Jianfei Guo, Xiaogang Jia, Yu~Deng, Zechu Li, Han Liu, Weiran Liao,
  Vignesh Prasad, Mathias Franzius, Gerhard Neumann, et~al.
\newblock Nautilus: From one prompt to plug-and-play robot learning.
\newblock \emph{arXiv preprint arXiv:2605.11665}, 2026.

\bibitem[Liang et~al.(2023)Liang, Huang, Xia, Xu, Hausman, Ichter, Florence,
  and Zeng]{liang2023code}
Jacky Liang, Wenlong Huang, Fei Xia, Peng Xu, Karol Hausman, Brian Ichter, Pete
  Florence, and Andy Zeng.
\newblock Code as policies: Language model programs for embodied control.
\newblock In \emph{2023 IEEE International conference on robotics and
  automation (ICRA)}, pp.\  9493--9500. IEEE, 2023.

\bibitem[Liu et~al.(2023)Liu, Zhu, Gao, Feng, Liu, Zhu, and
  Stone]{liu2023libero}
Bo~Liu, Yifeng Zhu, Chongkai Gao, Yihao Feng, Qiang Liu, Yuke Zhu, and Peter
  Stone.
\newblock Libero: Benchmarking knowledge transfer for lifelong robot learning.
\newblock \emph{Advances in Neural Information Processing Systems},
  36:\penalty0 44776--44791, 2023.

\bibitem[Ma et~al.(2024)Ma, Liang, Wang, Huang, Bastani, Jayaraman, Zhu, Fan,
  et~al.]{ma2024eureka}
Yecheng~Jason Ma, William Liang, Guanzhi Wang, De-An Huang, Osbert Bastani,
  Dinesh Jayaraman, Yuke Zhu, Jim Fan, et~al.
\newblock Eureka: Human-level reward design via coding large language models.
\newblock In \emph{International conference on learning Representations},
  volume 2024, pp.\  26516--26560, 2024.

\bibitem[Mees et~al.(2022)Mees, Hermann, Rosete-Beas, and
  Burgard]{mees2022calvin}
Oier Mees, Lukas Hermann, Erick Rosete-Beas, and Wolfram Burgard.
\newblock Calvin: A benchmark for language-conditioned policy learning for
  long-horizon robot manipulation tasks.
\newblock \emph{IEEE Robotics and Automation Letters}, 7\penalty0 (3):\penalty0
  7327--7334, 2022.

\bibitem[Merrill et~al.(2026)Merrill, Shaw, Carlini, Li, Raj, Bercovich, Shi,
  Shin, Walshe, Buchanan, et~al.]{merrill2026terminal}
Mike Merrill, Alexander Shaw, Nicholas Carlini, Boxuan Li, Harsh Raj, Ivan
  Bercovich, Lin Shi, Jeong Shin, Thomas Walshe, E~Kelly Buchanan, et~al.
\newblock Terminal-bench: Benchmarking agents on hard, realistic tasks in
  command line interfaces.
\newblock In \emph{International Conference on Learning Representations},
  volume 2026, pp.\  40903--40986, 2026.

\bibitem[Mittal et~al.(2025)Mittal, Roth, Tigue, Richard, Zhang, Du,
  Serrano-Munoz, Yao, Zurbr{\"u}gg, Rudin, et~al.]{mittal2025isaac}
Mayank Mittal, Pascal Roth, James Tigue, Antoine Richard, Octi Zhang, Peter Du,
  Antonio Serrano-Munoz, Xinjie Yao, Ren{\'e} Zurbr{\"u}gg, Nikita Rudin,
  et~al.
\newblock Isaac lab: A gpu-accelerated simulation framework for multi-modal
  robot learning.
\newblock \emph{arXiv preprint arXiv:2511.04831}, 2025.

\bibitem[Mu et~al.(2025)Mu, Chen, Chen, Peng, Lan, Gao, Liang, Yu, Zou, Xu,
  et~al.]{mu2025robotwin}
Yao Mu, Tianxing Chen, Zanxin Chen, Shijia Peng, Zhiqian Lan, Zeyu Gao, Zhixuan
  Liang, Qiaojun Yu, Yude Zou, Mingkun Xu, et~al.
\newblock Robotwin: Dual-arm robot benchmark with generative digital twins.
\newblock In \emph{2025 IEEE/CVF Conference on Computer Vision and Pattern
  Recognition (CVPR)}, pp.\  27649--27660. IEEE, 2025.

\bibitem[Nasiriany et~al.(2024)Nasiriany, Maddukuri, Zhang, Parikh, Lo, Joshi,
  Mandlekar, and Zhu]{nasiriany2024robocasa}
Soroush Nasiriany, Abhiram Maddukuri, Lance Zhang, Adeet Parikh, Aaron Lo,
  Abhishek Joshi, Ajay Mandlekar, and Yuke Zhu.
\newblock Robocasa: Large-scale simulation of everyday tasks for generalist
  robots.
\newblock \emph{arXiv preprint arXiv:2406.02523}, 2024.

\bibitem[{OpenAI}(2026{\natexlab{a}})]{openai2026gpt56}
{OpenAI}.
\newblock {GPT-5.6} system card, July 2026{\natexlab{a}}.
\newblock URL \url{https://deploymentsafety.openai.com/gpt-5-6}.

\bibitem[{OpenAI}(2026{\natexlab{b}})]{openai2026gpt6astra}
{OpenAI}.
\newblock {GPT-6 Astra} system card, September 2026{\natexlab{b}}.
\newblock URL \url{https://deploymentsafety.openai.com/gpt-6-astra}.

\bibitem[Qiang et~al.(2025)Qiang, Zhuang, Li, Sagar V~K, Zhang, Li, Wong, Yang,
  Liang, Zhang, and Dai]{qiang2026mle}
Rushi Qiang, Yuchen Zhuang, Yinghao Li, Dingu Sagar V~K, Rongzhi Zhang,
  ChangHao Li, Ian Wong, Sherry Yang, Percy Liang, Chao Zhang, and Bo~Dai.
\newblock Mle-dojo: Interactive environments for empowering llm agents in
  machine learning engineering.
\newblock In D.~Belgrave, C.~Zhang, H.~Lin, R.~Pascanu, P.~Koniusz,
  M.~Ghassemi, and N.~Chen (eds.), \emph{Advances in Neural Information
  Processing Systems}, volume 38, Main Conference. Curran Associates, Inc.,
  2025.
\newblock \doi{10.52202/085713-0139}.
\newblock URL
  \url{https://proceedings.neurips.cc/paper_files/paper/2025/file/0603c69125ad4b964bc9c4832f7b9f8f-Paper-Datasets_and_Benchmarks_Track.pdf}.

\bibitem[Sundermeyer et~al.(2021)Sundermeyer, Mousavian, Triebel, and
  Fox]{sundermeyer2021contact}
Martin Sundermeyer, Arsalan Mousavian, Rudolph Triebel, and Dieter Fox.
\newblock Contact-graspnet: Efficient 6-dof grasp generation in cluttered
  scenes.
\newblock In \emph{2021 IEEE international conference on robotics and
  automation (ICRA)}, pp.\  13438--13444. IEEE, 2021.

\bibitem[Tao et~al.(2024)Tao, Xiang, Shukla, Qin, Hinrichsen, Yuan, Bao, Lin,
  Liu, Chan, et~al.]{tao2024maniskill3}
Stone Tao, Fanbo Xiang, Arth Shukla, Yuzhe Qin, Xander Hinrichsen, Xiaodi Yuan,
  Chen Bao, Xinsong Lin, Yulin Liu, Tse-kai Chan, et~al.
\newblock Maniskill3: Gpu parallelized robotics simulation and rendering for
  generalizable embodied ai.
\newblock \emph{arXiv preprint arXiv:2410.00425}, 2024.

\bibitem[Tassa et~al.(2018)Tassa, Doron, Muldal, Erez, Li, Casas, Budden,
  Abdolmaleki, Merel, Lefrancq, et~al.]{tassa2018deepmind}
Yuval Tassa, Yotam Doron, Alistair Muldal, Tom Erez, Yazhe Li, Diego de~Las
  Casas, David Budden, Abbas Abdolmaleki, Josh Merel, Andrew Lefrancq, et~al.
\newblock Deepmind control suite.
\newblock \emph{arXiv preprint arXiv:1801.00690}, 2018.

\bibitem[Team(2026)]{team2026harbor}
Harbor~Framework Team.
\newblock Harbor: A framework for evaluating and optimizing agents and models
  in container environments.
\newblock \emph{Zenodo}, 2026.

\bibitem[Wu et~al.(2024)Wu, Shentu, Yi, Lin, and Abbeel]{wu2024gello}
Philipp Wu, Yide Shentu, Zhongke Yi, Xingyu Lin, and Pieter Abbeel.
\newblock Gello: A general, low-cost, and intuitive teleoperation framework for
  robot manipulators.
\newblock In \emph{2024 IEEE/RSJ International Conference on Intelligent Robots
  and Systems (IROS)}, pp.\  12156--12163. IEEE, 2024.

\bibitem[Xiao et~al.(2026)Xiao, Xie, Zhang, Lin, Fu, Xue, Lu, Yang, Dai, Wang,
  et~al.]{xiao2026enpire}
Wenli Xiao, Jia Xie, Tonghe Zhang, Haotian Lin, Letian Fu, Haoru Xue, Jalen Lu,
  Yi~Yang, Cunxi Dai, Zi~Wang, et~al.
\newblock Enpire: Agentic robot policy self-improvement in the real world.
\newblock \emph{arXiv preprint arXiv:2606.19980}, 2026.

\bibitem[Xie et~al.(2024{\natexlab{a}})Xie, Zhang, Chen, Li, Zhao, Cao, Hua,
  Cheng, Shin, Lei, et~al.]{xie2024osworld}
Tianbao Xie, Danyang Zhang, Jixuan Chen, Xiaochuan Li, Siheng Zhao, Ruisheng
  Cao, Toh~J Hua, Zhoujun Cheng, Dongchan Shin, Fangyu Lei, et~al.
\newblock Osworld: Benchmarking multimodal agents for open-ended tasks in real
  computer environments.
\newblock \emph{Advances in Neural Information Processing Systems},
  37:\penalty0 52040--52094, 2024{\natexlab{a}}.

\bibitem[Xie et~al.(2024{\natexlab{b}})Xie, Zhao, Wu, Liu, Luo, Zhong, Yang,
  and Yu]{xie2024text2reward}
Tianbao Xie, Siheng Zhao, Chen Wu, Yitao Liu, Qian Luo, Victor Zhong, Yanchao
  Yang, and Tao Yu.
\newblock Text2reward: Reward shaping with language models for reinforcement
  learning.
\newblock In \emph{International Conference on Learning Representations},
  volume 2024, pp.\  35663--35699, 2024{\natexlab{b}}.

\bibitem[Ye et~al.(2026)Ye, Ge, Zheng, Gao, Yu, Kurian, Indupuru, Tan, Zhu,
  Xiang, et~al.]{ye2026world}
Seonghyeon Ye, Yunhao Ge, Kaiyuan Zheng, Shenyuan Gao, Sihyun Yu, George
  Kurian, Suneel Indupuru, You~Liang Tan, Chuning Zhu, Jiannan Xiang, et~al.
\newblock World action models are zero-shot policies.
\newblock \emph{arXiv preprint arXiv:2602.15922}, 2026.

\bibitem[{Z.ai}(2026)]{zai2026glm53flash}
{Z.ai}.
\newblock {GLM-5.3-Flash}.
\newblock Hugging Face model card, 2026.
\newblock URL \url{https://huggingface.co/zai-org/GLM-5.3-Flash}.

\bibitem[Zhu et~al.(2020)Zhu, Wong, Mandlekar, Mart{\'\i}n-Mart{\'\i}n, Joshi,
  Lin, Maddukuri, Nasiriany, and Zhu]{zhu2020robosuite}
Yuke Zhu, Josiah Wong, Ajay Mandlekar, Roberto Mart{\'\i}n-Mart{\'\i}n,
  Abhishek Joshi, Kevin Lin, Abhiram Maddukuri, Soroush Nasiriany, and Yifeng
  Zhu.
\newblock robosuite: A modular simulation framework and benchmark for robot
  learning.
\newblock \emph{arXiv preprint arXiv:2009.12293}, 2020.

\end{thebibliography}
